\documentclass[final]{cvpr}

\usepackage{times}
\usepackage{epsfig}
\usepackage{graphicx}
\usepackage{array}
\usepackage{amsmath}
\usepackage{amssymb}
\usepackage{amsfonts}
\usepackage{booktabs}

\usepackage{caption}
\usepackage{subcaption}
\usepackage{graphicx}
\usepackage[hyphens]{url}
\usepackage{subfiles}

\usepackage{mathtools}

\usepackage{multirow}

\usepackage{algorithm}
\usepackage{algorithmic}
\usepackage{float}  
\usepackage{lipsum}


\DeclareMathOperator*{\E}{\mathbb{E}}

\usepackage[pagebackref=true,breaklinks=true,colorlinks,bookmarks=false]{hyperref}

\def\cvprPaperID{6877} 
\def\confYear{CVPR 2021}

\begin{document}

\title{Ternary Hashing}


\author{Chang Liu\\
Xi'an Jiaotong-Liverpool University\\
Suzhou\\
{\tt\small chang.liu17@student.xjtlu.edu.cn}
\and
Lixin Fan\\
WeBank AI\\
Shenzhen\\
{\tt\small lixinfan@webank.com}
\and
Kam Woh Ng\\
WeBank AI\\
Shenzhen\\
{\tt\small jinhewu@webank.com}
\and
Yilun Jin\\
The Hong Kong University of Science and Technology\\
Hong Kong\\
{\tt\small yilun.jin@connect.ust.hk}
\and
Ce Ju\\
WeBank AI\\
Shenzhen\\
{\tt\small ceju@webank.com}
\and
Tianyu Zhang\\
WeBank AI\\
Shenzhen\\
{\tt\small brutuszhang@webank.com}
\and
Chee Seng Chan\\
University of Malaya\\
Kuala Lumpur\\
{\tt\small cs.chan@um.edu.my}
\and
Qiang Yang\\
The Hong Kong University of Science and Technology\\
Hong Kong\\
{\tt\small qyang@cse.ust.hk}
}

\maketitle

\begin{abstract}

This paper proposes a novel ternary hash encoding for learning to hash methods, which provides a principled more efficient coding scheme with performances better than those of the state-of-the-art binary hashing counterparts. Two kinds of axiomatic ternary logic, Kleene logic and Łukasiewicz logic are adopted to calculate the Ternary Hamming Distance (THD) for both the learning/encoding and testing/querying phases. Our work demonstrates that, with an efficient implementation of ternary logic on standard binary machines, the proposed ternary hashing is compared favorably to the binary hashing methods with consistent improvements of retrieval mean average precision (mAP) ranging from 1\% to 5.9\% as shown in CIFAR10, NUS-WIDE and ImageNet100 datasets. 

\end{abstract}

\section{Introduction}

    
    Binary hashing has been proven as an efficient information retrieval technique with two advantages: (1) hash codes take less memory than the original data items (e.g. image), and (2) calculating the Hamming distance among hash codes is more efficient than calculating the Euclidean distance among original information. While a plethora of binary hashing methods have demonstrated noticeable performance improvements by taking the advantages of advances in recent deep neural networks (\cite{wang_pami_hashsurvey_18, luo2020survey}), one may wonder whether there is a principled approach to further boost performances of the state-of-the arts. This paper proposes such a novel method, \emph{ternary hashing}, which exploits ternary hash encoding to improve \emph{code efficiency} and \emph{retrieval accuracy} of existing binary hashing methods. 
    
    From the perspective of information theory, the most efficient encoding scheme should be based on the Euler's number $e$ \cite{third_base_hayes}, while 
    ternary encoding is arguably the most favorable natural number based encoding which admits practical implementations. Moreover,  a third state UNKNOWN was introduced in \emph{ternary logic} to represent the ambiguous states between TRUE and FALSE \cite{RTLF_Masao86,bvelohlavek2017fuzzy}. When applying ternary logic to hashing, the proposed Ternary Hamming Distance (THD) essentially assigns smaller weights to ambiguous cases involving UNKNOWN points and thus reduces the ambiguity in Hamming distance measures (see Sect. \ref{subsect:CodeError}). 
    
    Despite the two aforementioned advantages, a number of open questions must be properly addressed to exploit the efficacy of ternary hashing. First, how to implement ternary hashing on the widely available binary machines? Second, how to ensure that such an implementation still enjoys the benefits provided by ternary encoding?  Third, how to fully exploit the advantageous ternary hashing with true ternary machines? In this paper, we provide theoretical and empirical justifications to answer the first two questions (Sect. \ref{sect:ternary-hashing}), while discuss possible solutions to the third question e.g. with FPGA implementation for future work (Sect. \ref{sect-dis-conl}).

    Our contribution are three-folds: (1) we propose a novel ternary hashing method to exploit the efficiency of ternary coding. To our best knowledge, this is the first work that investigates different \textit{ternary logic} in the context of Learning to Hashing (Sect. \ref{sect:ternary-hashing}); (2) we provide both theoretical and empirical justifications of the advantageous ternary hashing, in terms of its ability to \emph{reduce neighborhood ambiguity} (Sect. \ref{subsect:reduce-amb-by-th} and \ref{subsect:exper-perf});  and (3) we give an efficient implementation of ternary hashing on binary machines, showing that ternary hashing is a feasible and preferable solution for learning to hash (LtH) even on the widely available binary machines (Sect. \ref{Sect:TernaryImple.}).

\section{Related Work}

	\begin{table*}[!htb]
	\begin{minipage}{.33\linewidth}
		\centering
		\resizebox{\linewidth}{!}{
			\begin{tabular}{|c|c|ccc|ccc|ccc|c|}
				\hline
				\multicolumn{2}{|c|}{}  & \multicolumn{3}{c|}{$A \wedge B$} & \multicolumn{3}{c|}{$A \vee B$} & \multicolumn{3}{c|}{$A \leftrightarrow B$}  & $\neg A$\\ \hline
				\multicolumn{2}{|c|}{B} & \textbf{-1}        & \textbf{0}         & \textbf{+1}        & \textbf{-1}        & \textbf{0}        & \textbf{+1}       & \textbf{-1}           & \textbf{0}            & \textbf{+1}    &      \\ \hline
				\multirow{3}{*}{A} & \textbf{-1} & -1        & -1        & -1        & -1        & 0        & +1       & +1           & 0            & -1      & +1     \\ \cline{2-12} 
				& \textbf{0}  & -1        & 0         & 0         & 0         & 0        & +1       & 0            & +1           & 0      & 0      \\ \cline{2-12} 
				& \textbf{+1} & -1        & 0         & +1        & +1        & +1       & +1       & -1           & 0            & +1     & -1      \\ \hline
		\end{tabular}}
		\caption{Łukasiewicz Logic \label{tab:lukas}}
	\end{minipage}%
	\hfill
	\begin{minipage}{.33\linewidth}
		\centering
		\resizebox{\linewidth}{!}{
			\begin{tabular}{|c|c|ccc|ccc|ccc|c|}
				\hline
				\multicolumn{2}{|c|}{}  & \multicolumn{3}{c|}{$A \wedge B$} & \multicolumn{3}{c|}{$A \vee B$} & \multicolumn{3}{c|}{$A \leftrightarrow B$} 
				& $\neg A$\\ \hline
				\multicolumn{2}{|c|}{B} & \textbf{-1}        & \textbf{0}         & \textbf{+1}        & \textbf{-1}        & \textbf{0}        & \textbf{+1}       & \textbf{-1}           & \textbf{0}            & \textbf{+1}     &      \\ \hline
				\multirow{3}{*}{A} & \textbf{-1} & -1        & -1        & -1        & -1        & 0        & +1       & +1            & 0           & -1        & +1   \\ \cline{2-12} 
				& \textbf{0}  & -1        & 0         & 0         & 0         & 0        & +1       & 0             & 0           & 0       & 0     \\ \cline{2-12} 
				& \textbf{+1} & -1        & 0         & +1        & +1        & +1       & +1       & -1            & 0           & +1     & -1      \\ \hline
		\end{tabular}}
		\caption{Kleene Logic \label{tab:Kleen}}
	\end{minipage} 
	\hfill
	\begin{minipage}{.33\linewidth}
		\centering
		\resizebox{\linewidth}{!}{
			\begin{tabular}{|c|c|ccc|ccc|ccc|c|}
				\hline
				\multicolumn{2}{|c|}{}  & \multicolumn{3}{c|}{$A \wedge B$} & \multicolumn{3}{c|}{$A \vee B$} & \multicolumn{3}{c|}{$A \leftrightarrow B$} & $\neg A$ \\ \hline
				\multicolumn{2}{|c|}{B} & \textbf{-1}        & \textbf{0}         & \textbf{+1}        & \textbf{-1}        & \textbf{0}        & \textbf{+1}       & \textbf{-1}           & \textbf{0}            & \textbf{+1}     &        \\ \hline
				\multirow{3}{*}{A} & \textbf{-1} & -1         & 0        & -1        & -1        & 0        & +1       & +1            & 0           & -1      & +1     \\ \cline{2-12} 
				& \textbf{0}  & 0          & 0        & 0         & 0         & 0        & 0        & 0             & 0           & 0     & 0       \\ \cline{2-12} 
				& \textbf{+1} & -1         & 0        & +1        & +1        & 0        & +1       & -1            & 0           & +1    & -1       \\ \hline
		\end{tabular}}
		\caption{Bochvar Logic \label{tab:Bochvar}}
	\end{minipage} 
\end{table*}


	\subsection{Binary Hashing}
	
	Existing hashing methods can be roughly divided into two categories: unsupervised hashing and supervised hashing. Unsupervised hashing methods usually learn a hash function to map data into binary codes without labeled data. Locality Sensitive Hashing (LSH) \cite{gionis1999similarity} aims to maximize the probability that similar data are mapped into similar binary codes. Spectral Hashing \cite{yair_nips_sh_09} compute binary codes via Principal Components Analysis (PCA) on the data. Supervised hashing methods learn a hash function with labeled data. As learning to map data into binary codes directly will cause ill-posed gradient problem, HashNet \cite{HashNet_CaoICCV2017} was proposed to optimize the Deep Neural Network (DNN) with continuation method. GreedyHash \cite{GreedyHash_SuNIPS18} designed a hash coding layer that uses the sign function in forward propagation to maintain the discrete constraints and the gradients are transmitted directly to front layer to avoid the vanishing gradient. JMLH \cite{JMLH_ShenICCV19w} employs Information Bottleneck (IB) by minimizing the Kullback-Leibler (KL) divergence between posterior hash codes $q(b|x)$ and the code prior $p(b)$ through a classification model. MMHH \cite{MaxMarginHash_KangICCV19} introduces a max-margin t-distribution loss to concentrate more similar data points to be within a Hamming ball, and the margin is a penalization to the Hamming radius.
	For a more comprehensive survey, please refer to \cite{wang_pami_hashsurvey_18, luo2020survey}. 

	\subsection{Ternary Assignment}
	
	One crux of binary hashing is the wrong assignments of ambiguous data values into either 0 or 1, and inevitably resulting in the loss of retrieval accuracy.  This issue is well recognized, as\textit{ quantization error}, and methods have been proposed to mitigate the error e.g. by {assigning ambiguous data values into double-bit binary codes with Hamming distances smaller than those of clear-cut data values} \cite{kong2012double}.  This heuristic double-bit assignment, in our view, is linked to the implementation of Łukasiewicz ternary logic on binary machines. 
	However, its underlying ternary logic nature was not disclosed in \cite{kong2012double}. 

	\cite{fandeep} proposed Deep Polarized Network (DPN) and a hinge-like polarization loss to "polarize" the output $z$. The polarized network outputs z was converted into ternary codes for image retrieval, where the values of $z$ between double thresholds $(-m, m)$ are assigned as 0, others are assigned as $1$ or $-1$ according to their sign. However, the setting of thresholds $(-m, m)$ in \cite{fandeep} was somehow arbitrary and sub-optimal, as disclosed by investigations in our paper (Sect. \ref{sect:ternary-hashing}).  
	
	
	
	\subsection{Ternary Quantization \& Hardware}
	A number of ternary neural networks were proposed to take advantages of efficient ternary coding and have demonstrated significant speed gains and reduced network complexity \cite{wei_nips_terngrad_17,alemdar_ijcnn_tnn_17,zhang2019neural}.  However, these methods only quantize network weights into three values and network outputs are still kept as real-valued variables. 
	Recent studies on hardware implementation of ternary neural network \cite{alemdar_ijcnn_tnn_17, tridgell_acm_utnn_19, boucle_fpl_fpga_17} also shown that it is possible to have an efficient and fast ternary-based computation on Field-Programmable Gate Array (FPGA) \cite{stephen_fpga_92}.

	




\section{Ternary Hashing} \label{sect:ternary-hashing}

\begin{figure*}
	\centering
	\includegraphics[width=\linewidth]{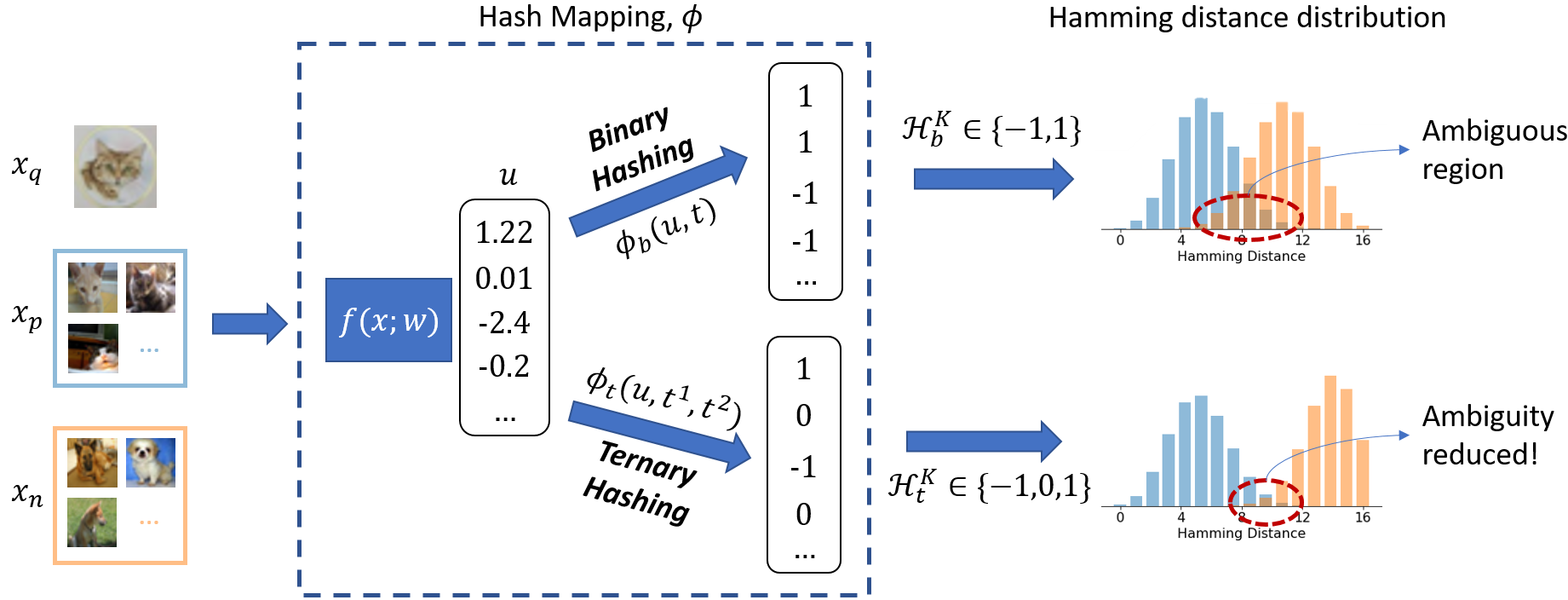}
	\caption{Overview of ternary hashing. \textbf{(Left)} Given the query image $x_q$, positive images $x_p$ and negative images $x_n$, \textbf{(Middle)} the respective hash codes $\mathcal{H}^K$ are generated by hash mapping function $\phi(f(x; w), t)$, where $t$ is either \textit{bias terms} in the binary hashing network or $t=(t^1,t^2)$ the double threshold found by the proposed ternary hashing method (see Algorithm \ref{alg:find_mu}). \textbf{(Right)} Hamming distance distributions are yield by computing hamming distances $d_\phi$ between the hash code of the query image and the hash code of both positive and negative images. Ternary hamming distance distribution has \textit{better separability} than binary hamming distance distribution due to the reduced probability of bit assignment errors.
		\label{fig:comp-hist}}
\end{figure*}

This section describes a novel post-processing ternary hash encoding for binary hash codes. We first formulate ternary logic with Łukasiewicz, Kleene and Bochvac logic in Sect. \ref{subsect:ternary-logic} and explain neighborhood ambiguity by hashing in Sect. \ref{subsect:CodeError}. Next, we describe our proposed algorithm that is to reduce neighborhood ambiguity of hash codes in Sect. \ref{subsect:reduce-amb-by-th} and finally we illustrate how to implement a ternary hash codes on a binary machine in Sect. \ref{Sect:TernaryImple.}

\subsection{Ternary Logic} \label{subsect:ternary-logic}

The study of different forms of ternary logic were initiated by Łukasiewicz, Kleene and Bochvac etc. and the first axiomatic ternary logic was ascribed to Moisil \cite{bvelohlavek2017fuzzy}. Similar to their binary logic counterparts, logic operators like AND ($\wedge$), OR ($\vee$), EQUIVALENCE ($\leftrightarrow$), XOR ($\oplus$), NOT($\neg$) etc. were well studied. 
For the sake of brevity, we only review below these logic operators that are used in our work and refer to \cite{bvelohlavek2017fuzzy} for a thorough treatment of fuzzy \& ternary logic.

Tables \ref{tab:lukas}, \ref{tab:Kleen} and \ref{tab:Bochvar} illustrate truth tables of four logic operators AND  ($\wedge$), OR ($\vee$), EQUIVALENCE ($\leftrightarrow$) and NOT ($\neg$), respectively, for Łukasiewicz, Kleene and Bochvac logic. One must note that, an UNKNOWN state denoted as $0$ is assigned between TRUE and FALSE denoted as $+1, -1$. As shown below this UNKNOWN state plays crucial role in dealing with ambiguous data points.   Also one may note that, for both Kleene and Bochvac logic, truth tables of $A \leftrightarrow B$ are the same and $0 \leftrightarrow 0 = 0$, while for Łukasiewicz logic, there is a subtle difference i.e. $0 \leftrightarrow 0 = +1$. This difference leads to slightly higher hashing accuracies  for Kleene and Bochvac logic (see Sect. \ref{sect-expr}). 

For two ternary codes or \textit{trits} $a,b \in \{-1, 0, +1\}$, we propose to adopt \textit{ternary Hamming distance} $\mathsf{THD}(a, b)$ as follows,
\begin{equation}\label{ternaryHamm}
  \mathsf{THD}(a, b) := \frac{1}{2}( \neg(a \leftrightarrow b) + 1 ) \rightarrow \{0, 0.5, 1\}, 
\end{equation}
in which operators $\neg, \leftrightarrow$ are given in Tables \ref{tab:lukas}, \ref{tab:Kleen} and \ref{tab:Bochvar}.  

Note that for normal binary bits taking values $\{-1, +1\}$ as inputs the proposed $\mathsf{THD}$ is de-generalized as the binary hamming distance, which assigns distances either 0 or 1 between two bits.  However, a smaller distance of $0.5$ is assigned to two trits  taking at least one middle value 0 (except for Łukasiewicz logic which assigns distance 1 between two middle valued trits).  This moderate distance measurement for middle-valued data points turns out to be crucial in dealing with ambiguity of hamming distance measurements. 


\subsection{Neighborhood Ambiguity by Hashing} 
\label{subsect:CodeError}

The motivation of ternary hashing bears some similarities to \cite{MI_Hash_PAMI19} which proposed to \emph{minimize neighborhood ambiguity} in the learned Hamming space.  We show below that the neighborhood ambiguity is actually rooted in the probability of mis-classifications for assigning wrong codes to each bit. This error is inevitable for inseparable cases and  is lower bounded by the \textit{Bayes error} even if an optimal decision was made  \cite{JSDHash_FanICCV13,LinJSD91}. Within this formulation, the conditional Hamming distance distribution (i.e. $D_{\hat{x}, \Phi}$ in \cite{MI_Hash_PAMI19}) follows the \emph{Poisson binomial distribution} with independent Bernoulli distributions parameterized by varying error rates in assigning each bit. Moreover, it was shown that Ternary Hashing helps to reduce errors in each bit assignments by replacing a binary threshold with a pair of double thresholds defining the ambiguous region (see Figure \ref{fig:visualize_dist}). 
Consequently, Ternary Hamming Distance leads to less ambiguous Hamming distance measures as compared with binary Hamming distance counterparts (see Figure \ref{fig:comp-hist} and \ref{fig:ambiguity}). 


Let $\mathcal{X} \subset \mathbb{R}^N$ be the feature space and $\mathcal{H}^K := \{-1,+1\}^K$ be the Hamming space, following \cite{MI_Hash_PAMI19,fandeep}, the ultimate goal of supervised learning to hash (LtH) is to learn a similarity-preserving Hamming embedding function $\Phi : \mathcal{X} \rightarrow \mathcal{H}^K $ that satisfies the following requirement: 
\begin{equation}
    d_{\Phi}(x_q, x_p) < d_{\Phi}(x_q, x_n), \label{hamdistreq}
\end{equation}
in which $x_p, x_n \in \mathcal{X}$ are, respectively, positive/negative points to the \textit{query} $x_q \in \mathcal{X}$, and  $\Phi = \{\phi_1, \cdots, \phi_K\}$ is a set of $K$ mappings $\phi_i(x)=\text{sgn}(u_i) \in \{-1, +1\}$, where $u_i=f_i(x; w_i, t_i)$, which are often parameterized by channel-wise network weights $w_i$ and thresholds (or bias terms) $t_i$. 



The requirement (\ref{hamdistreq}) might not be fulfilled and the amount of violation is considered as the degree of neighborhood ambiguity in \cite{MI_Hash_PAMI19}. 
Let $\mathsf{Dm} := d_{\Phi}(x_q, x_n)  - d_{\Phi}(x_q, x_p)$ quantify the amount of ambiguity, such that large \textit{negative} values in $\mathsf{Dm}$ indicate  highly ambiguous cases and large \textit{positive} values for separable cases. By re-arranging (\ref{hamdistreq}) and the triangle inequality, it follows that 
\begin{equation}
  | d_{\Phi}(x_p, x_n) - \mathsf{Dm} | < 2 d_{\Phi}(x_q, x_p). \label{hamdistreq-bound}
\end{equation}
Note that $\mathsf{Dm}$ is closely related to $ d_{\Phi}(x_p, x_n)$ since the difference in between is bounded by $2d_{\Phi}(x_q, x_p)$ which are small non-negative values for points $x_p$ similar to the query $x_q$. One may therefore push $\mathsf{Dm}$ toward positive regime and reduce the ambiguity, by maximizing the expectation of $d_{\Phi}(x_p, x_n)$ over sets of all positive and negative examples $x_p, x_n$,  
\begin{equation}
\Phi^* := \arg \max_{\Phi} \E\limits_{x_p \in \mathbf{X}_p, \\x_n \in \mathbf{X}_n}( d_{\Phi}(x_p, x_n)). \\
\end{equation}





Due to the additive nature of Hamming distance $d_{\Phi}(x_p, x_n) = \sum^K_{i=1} d_{\phi_i}(x_p, x_n)$, one may seek to maximize the Hamming distance in a bitwise manner, i.e. 
\begin{equation}
(w_i, t_i)^*  = \arg \max_{w_i, t_i} \E\limits_{x_p \in \mathbf{X}_p, \\x_n \in \mathbf{X}_n}( d_{\phi_i}(x_p, x_n)), 
\end{equation}
\begin{equation}
d_{\phi_i}(x_p, x_n) = \left\{ \begin{array}{ll} 
        0 & \mbox{if ${\phi_i}(x_p) = {\phi_i}(x_n)$};\\
        1 & \mbox{if ${\phi_i}(x_p) \neq {\phi_i}(x_n)$}.\end{array}  \right.
\end{equation}
in which $\E(d_{\phi_i}(x_p, x_n)) = p_i \cdot 1 + (1-p_i) \cdot 0 = p_i$ essentially quantifies the probability $p_i$ of successfully distinguishing  positive from negative samples  $x_p, x_n$. For inseparable distributions, 
the probability of mis-classifications errors $\Bar{p}_i = 1-p_i$ is non-negligible. 
In fact, this probability of error is lower bounded by the \textit{Bayes error} in case an optimal decision was made \cite{JSDHash_FanICCV13,LinJSD91}. 


Formally, let $\mathcal{B}_i := ( d_{\phi_i}(x_p, x_n)) \sim \text{Bernoulli}(p_i), i=1,\cdots K$ be a set of $K$ independent and \emph{non-identically} distributed Bernoulli indicators. Here we assume both positive and negative samples are independently drew from corresponding sets i.e. $x_p \in \mathbf{X}_p, x_n \in \mathbf{X}_n$ and  $p_i=Pr(\mathcal{B}_i=1)$ is the probability of positive and negative samples being successfully classified by each indicator. The sum of random indicators $d_{\Phi}(x_p, x_n) = \sum^K_{i=1} \mathcal{B}_i$ is the \textit{Poisson binomial} random variable taking values in $\{0,\cdots K\}$.  Given the error probabilities for $\mathcal{B}_i$, it is thus possible to compute the conditional Hamming distance with a numerical method dedicated for the Poisson binomial distribution \cite{PBD_HONG13}.  This capability allows us to numerically study the relationship between the ambiguity in Hamming distance measurements and the probability of errors in bit assignments brought by any hashing methods. 
It is empirically justified that \textit{reducing the probability of error in bit assignments leads to reduced ambiguity} in Hamming distance measurements i.e. better separability between positive/negative data points (see Figure \ref{fig:ambiguity}).    

\subsection{Reducing Neighborhood Ambiguity By Ternary Hashing}
\label{subsect:reduce-amb-by-th}

In order to reduce neighborhood ambiguity, we propose to reduce the error probability $\Bar{p}_i$ by adopting the ternary logic and ternary Hamming distance introduced earlier. 
Specifically, for each ternary hashing  mapping $\phi^t_i( v_i, t^1_i, t^2_i) \in \{-1, 0, +1\} $ which  assigns ternary codes to network outputs $v_i=f_i(x; w_i)$ according to double thresholds $t^1_i, t^2_i$,   a \textbf{double-threshold searching algorithm} is adopted to find a pair of thresholds defining an ambiguous region in which data points are assigned with a middle value $0$ (see Algorithm \ref{alg:find_mu}).  

Following (\ref{ternaryHamm}) and Tables  \ref{tab:lukas}, \ref{tab:Kleen} and \
ref{tab:Bochvar},  the expectation of ternary Hamming distance is given by
\begin{equation}\label{eq:ETHD}
     \E(d^t_{\phi^t_i}(x_p, x_n)) = p_i \cdot 1 + u_i \cdot 0.5 + (1-p_i-u_i)  \cdot 0,  
\end{equation}
in which $d^t_{\phi^t_i}(x_p, x_n) = \mathsf{THD}( {\phi^t_i}(x_p), {\phi^t_i}(x_n) )$ measures the ternary hamming distance between network outputs for  positive and negative samples $ {\phi^t_i}(x_p), {\phi^t_i}(x_n)$, and $p_i, u_i$ are probabilities of ternary hamming distances $d^t_{\phi^t_i}(x_p, x_n)$ being assigned 1 and 0.5, respectively. 

Let $P^T_a, P^U_a, P^F_a$ denote the probabilities that a set of samples being assigned codes $+1, 0, -1$ respectively  (and  $P^T_b, P^U_b, P^F_b$ being defined in a similar vein for another set of samples), according to truth tables \ref{tab:lukas}, \ref{tab:Kleen} and \ref{tab:Bochvar}, the expectation of ternary hamming distances in (\ref{eq:ETHD}) is then given by, 
\begin{equation}
\begin{split}
\mathbb{E}_{L}(A, B) = & 1 \cdot \Big(P_a^T P_b^F + P_a^F P_b^T \Big)  + \\ & 0.5 \cdot \Big( P_a^U (P_b^T + P_b^F) +  P_b^U (P_a^T + P_a^F) \Big),
\label{eq:lukahd}
\end{split}
\end{equation}
\begin{equation}
\begin{split}
\mathbb{E}_{K}(A,B) = & 1 \cdot \Big(P_a^T P_b^F + P_a^F P_b^T \Big)  +  0.5 \cdot P_a^U P_b^U + \\ & 0.5 \cdot \Big( P_a^U (P_b^T + P_b^F) +  P_y^U (P_a^T + P_a^F) \Big),
\label{eq:kleenhd}
\end{split}
\end{equation}
in which $\mathbb{E}_{L}, \mathbb{E}_{K}$ are ternary hamming distances between two sets of samples $a \in {A}, b \in {B}$ for Łukasiewicz and Kleene/Bochavar logic, respectively.

Ternary hamming distances in (\ref{eq:lukahd}) and (\ref{eq:kleenhd}) are used to quantify (dis-)similarity between samples from two different classes. For samples belonging to $C$ different classes  $ X := X_1 \cup  \cdots \cup X_C$, the  pairwise ternary hamming distances over all possible pairs of sets is given by 
\begin{equation}\label{eq:pairTHD}
\mathbb{E}_{K,L}(X) =  \sum_{A, B \in \{X_1, \cdots, X_C\}} (-1)^s \mathbb{E}_{K,L} (A, B), 
\end{equation}
where $s=\Big \{ \begin{array}{lr}
0  & \text{if } A \neq B\\
1 &  \text{if } A=B
\end{array}$.  Note that $\mathbb{E}_{K,L}(X)$ are respective Kleene and  Łukasiewicz ternary hamming distances, and we may omit the subscript for brevity. 

Using (\ref{eq:pairTHD}),  we propose a double-threshold searching algorithm that finds, for each ternary hashing mapping, a pair of thresholds that \emph{maximize the expectation of pairwise ternary hamming distances between negative samples, minus ternary hamming distances between positive samples} (line 15 in Algorithm \ref{alg:find_mu}). 

\begin{algorithm}
   \caption{Double-threshold Searching}
   \label{alg:find_mu}  
   \begin{algorithmic}[1]
		\REQUIRE
			$v_i^A$: Raw network output of i-th bit of database A
			
			$v_i^B$: Raw network output of i-th bit of database B

			$R$: number of bins
			
			$C$: number of classes
			
		\ENSURE
			$t_1, t_2$: Best Threshold
		\STATE $A^{Hist} = \text{matrix of } C \times R $
		\STATE $B^{Hist} = \text{matrix of } C \times R$
		\STATE $d = \text{matrix of } R \times R$
		\STATE $A^{Hist} = histogram(u_i^A, C, R)$
		\STATE $B^{Hist} = histogram(u_i^B, C, R)$
		\STATE interval $=$ range of $A^{Hist}$ and $B^{Hist}$ \COMMENT{A vector of size $R$}
		\FOR{$i = 0 \to R$}
			\FOR{$j = i + 1 \to R$}
				\STATE $P_a^T = \sum_{k < i} A^{Hist}_{C,k}$ \COMMENT{$P$ is a vector of size $C$}
				\STATE $P_a^U = \sum_{i \le k \le j} A^{Hist}_{C,k}$
				\STATE $P_a^F = \sum_{k > j} A^{Hist}_{C,k}$ 
				\STATE $P_b^T = \sum_{k < i} B^{Hist}_{C,k}$
				\STATE $P_b^U = \sum_{i \le k \le j} B^{Hist}_{C,k}$
				\STATE $P_b^F = \sum_{k > j} B^{Hist}_{C,k}$
				\STATE compute $\mathbb{E}_{i,j}$ following eq. (\ref{eq:pairTHD}), 
			\ENDFOR
		\ENDFOR
		\STATE $i^{max}, j^{max} = \arg\max_{i,j}(\mathbb{E})$ 
		\STATE $t_1, t_2 = \text{interval}[i^{max}], \text{interval}[j^{max}]$
		\RETURN $t_1, t_2$
   \end{algorithmic}

\end{algorithm}

\subsection{Ternary Hashing on Binary Machine} \label{Sect:TernaryImple.}


In terms of coding efficiency per se, the most economical radix for a numbering system is $e$ whereas the product of the radix and the number of digits needed to express a given range is minimized \cite{third_base_hayes}. Since 3 is the integer closest to $e$, ternary coding is arguably the most efficient coding scheme that admits practical implementations (see discussions in Sect. \ref{sect-dis-conl}).

Nevertheless, binary coding systems are ubiquitous because  binary devices are simple and reliable, with just two stable states to encode.  In order to make it feasible to run the proposed ternary hashing on widely available binary machines, the representation of ternary coding is shown in Table \ref{tab:my_label} and ternary hamming distance is measured as below:

\begin{table}[t]
    \centering
    \begin{tabular}{|c|c|c|c|}
    \hline
    trit           & -1 & 0  & 1  \\ \hline
    $\overline{t_1t_2}$ & 01 & 00 & 10 \\ \hline
    \end{tabular}
    \caption{Binary representation of a trit with two bits $\overline{t_1 t_2}$.}
    \label{tab:my_label}
\end{table}

\noindent{\textbf{Łukasiewicz Hamming distance}}
\begin{equation}
\mathsf{THD}_l(a, b) = \frac{1}{2} \times \sum (\overline{t_1t_2} \oplus \overline{t_1't_2'})
\end{equation}

\noindent{\textbf{Kleene and Bochvar Hamming distance}}
\begin{equation}
	\begin{aligned}
		&& t_b & = \neg(t_1 \vee t_2 \vee t_1' \vee t_2') & \\
		&& \mathsf{THD}_k(a, b) & = \frac{1}{2} \times \sum ((\overline{t_1t_2} \oplus \overline{t_1't_2'}) \vee \overline{0t_b}) &
	\end{aligned}
\label{KleeneEQ}
\end{equation}
in which $a=\overline{t_1t_2}$ and $b=\overline{t_1't_2'}$ are representations of two trits on binary machines,  $\oplus$ denotes XOR operation and $\sum$ denotes counting the number of 1 in the binary value\footnote{Counting "1" bits can be efficiently implemented with \textit{popcnt} instruction. } 

Note that with this code representation,  '11' is actually not used, thus coding efficiency is only 75\% of the ideal case (see discussion in Sect. \ref{sect-dis-conl}). Nonetheless, even with this sub-optimal implementation, the ternary hashing can still outperform the binary hashing with same length of hash code. 
In terms of computational speed, the calculation of ternary Hamming distance is comparable with that of binary Hamming distance, and much faster than the floating-point number based matrix multiplication (Sect. \ref{sect-expr}).  

Interestingly, the implementation and the distance calculation of DBQ approach \cite{kong2012double} is the same as Łukasiewicz Hamming distance, while the distance calculation of DPN-T \cite{fandeep} is the same as Kleene Hamming distance. Performances differences between these two $\mathsf{THD}$ are discussed in Sect. \ref{sect-expr}. 

\section{Experiments}\label{sect-expr}


















Our proposed ternary hashing method (Algorithm \ref{alg:find_mu}) acted as a \textbf{post-processing} of the outputs of binary hashing network, and therefore it can be applied to \textit{any} binary hashing methods. In this section, we first describe the datasets and deep hashing methods that are used for comparison in this work (Sect. \ref{subsect:exper-set}), followed by performances of ternary hashing in terms of retrieval accuracy and efficiency (Sect. \ref{subsect:exper-perf}).  


\subsection{Experiment Settings}\label{subsect:exper-set}

\begin{table*}[h]
	\centering
	\begin{tabular}{l|c|c|c||c|c|c||c|c|c}
		\hline
		\multirow{2}{*}{Methods} & \multicolumn{3}{c||}{CIFAR-10} & \multicolumn{3}{c||}{NUS-WIDE} & \multicolumn{3}{c}{ImageNet100} \\ \cline{2-10}
		& \textbf{B}  & \textbf{L}  & \textbf{K} & \textbf{B}  & \textbf{L}  & \textbf{K} & \textbf{B}  & \textbf{L}  & \textbf{K} \\ \hline
		GH-16 & 0.815 & 0.820 & 0.825 & 0.730 & 0.786 & 0.768 & 0.584 & 0.609 & 0.621 \\
		GH-24 & 0.821 & 0.826 & 0.829 & 0.775 & 0.809 & 0.820 & 0.633 & 0.649 & 0.667 \\
		GH-32 & 0.823 & 0.827 & 0.830 & 0.788 & 0.820 & 0.834 & 0.653 & 0.666 & 0.677 \\
		GH-64 & 0.826 & 0.827 & 0.825 & 0.821 & 0.834 & 0.850 & 0.678 & 0.687 & 0.694 \\
		GH-128 & 0.811 & 0.811 & 0.811 & 0.827 & 0.838 & 0.854 & 0.676 & 0.682 & 0.688 \\
		\midrule
		JMLH-16 & 0.792 & 0.799 & 0.802 & 0.636 & 0.682 & 0.666 & 0.603 & 0.616 & 0.658 \\
		JMLH-24 & 0.798 & 0.810 & 0.814 & 0.671 & 0.699 & 0.698 & 0.632 & 0.632 & 0.670 \\
		JMLH-32 & 0.810 & 0.813 & 0.814 & 0.681 & 0.710 & 0.718 & 0.645 & 0.636 & 0.678 \\
		JMLH-64 & 0.819 & 0.823 & 0.824 & 0.717 & 0.731 & 0.755 & 0.658 & 0.661 & 0.688 \\
		JMLH-128 & 0.823 & 0.825 & 0.827 & 0.728 & 0.743 & 0.774 & 0.663 & 0.661 & 0.687 \\
		\midrule
		DPN-16 & 0.803 & 0.814 & 0.820 & 0.813 & 0.830 & 0.818 & \textbf{0.609} & \textbf{0.632} & \textbf{0.668} \\
		DPN-24 & 0.809 & 0.815 & 0.823 & 0.832 & 0.841 & 0.846 & 0.664 & 0.675 & 0.710 \\
		DPN-32 & 0.822 & 0.827 & 0.836 & 0.832 & 0.843 & 0.842 & 0.690 & 0.687 & 0.732 \\
		DPN-64 & 0.813 & 0.824 & 0.836 & 0.840 & 0.846 & 0.847 & 0.730 & 0.716 & 0.754 \\
		DPN-128 & 0.811 & 0.814 & 0.825 & 0.835 & 0.840 & 0.845 & 0.734 & 0.717 & 0.762 \\
		\midrule
		DPSH-16 & 0.798 & 0.803 & 0.811 & 0.830 & 0.831 & 0.824 & 0.293 & 0.310 & 0.318 \\
		DPSH-24 & 0.789 & 0.801 & 0.805 & 0.837 & 0.832 & 0.828 & 0.346 & 0.366 & 0.373 \\
		DPSH-32 & 0.793 & 0.800 & 0.806 & 0.843 & 0.836 & 0.834 & 0.376 & 0.397 & 0.409 \\
		DPSH-64 & 0.784 & 0.784 & 0.805 & 0.853 & 0.844 & 0.844 & 0.422 & 0.445 & 0.453 \\
		DPSH-128 & 0.753 & 0.725 & 0.766 & 0.861 & 0.845 & 0.851 & 0.436 & 0.454 & 0.465 \\
		\midrule
		HashNet-16 & 0.799 & 0.807 & 0.818 & 0.819 & 0.823 & 0.829 & 0.360 & 0.410 & 0.416 \\
		HashNet-24 & 0.788 & 0.811 & 0.821 & 0.834 & 0.828 & 0.838 & 0.432 & 0.474 & 0.482 \\
		HashNet-32 & 0.798 & 0.799 & 0.824 & 0.834 & 0.822 & 0.830 & 0.497 & 0.527 & 0.543 \\
		HashNet-64 & 0.789 & 0.764 & 0.811 & 0.846 & 0.816 & 0.844 & 0.587 & 0.607 & 0.618 \\
		HashNet-128 & 0.702 & 0.675 & 0.728 & 0.853 & 0.813 & 0.825 & 0.624 & 0.631 & 0.650 \\
		\midrule
		DBDH-16 & 0.804 & 0.810 & 0.818 & 0.820 & 0.829 & 0.823 & 0.253 & 0.286 & 0.290 \\
		DBDH-24 & 0.797 & 0.811 & 0.822 & 0.830 & 0.829 & 0.828 & 0.323 & 0.362 & 0.370 \\
		DBDH-32 & 0.793 & 0.802 & 0.817 & 0.840 & 0.837 & 0.839 & 0.355 & 0.393 & 0.409 \\
		DBDH-64 & 0.787 & 0.766 & 0.807 & 0.848 & 0.825 & 0.841 & 0.424 & 0.455 & 0.466 \\
		DBDH-128 & 0.764 & 0.780 & 0.800 & 0.855 & 0.830 & 0.854 & 0.475 & 0.492 & 0.494 \\
		\hline
	\end{tabular}
	\caption{Retrieval mAP of different methods with number of K in the suffix. B is the performance of binary hashing, while L and K are the performance of ternary hashing using Łukasiewicz logic or Kleene logic. The retrieval mAP are measured using bin size $R = 100$.}
	\label{tab:performance}
\end{table*}

\begin{figure*}[h]
	\centering
	\includegraphics[width=0.28\textwidth]{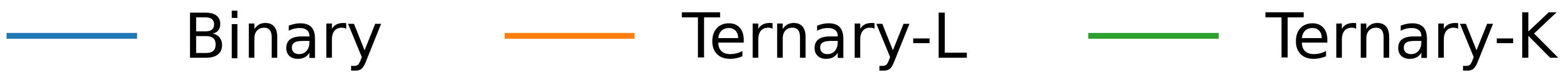}
	\\
	\begin{subfigure}[b]{0.28\textwidth}
		\centering
		\includegraphics[width=\textwidth]{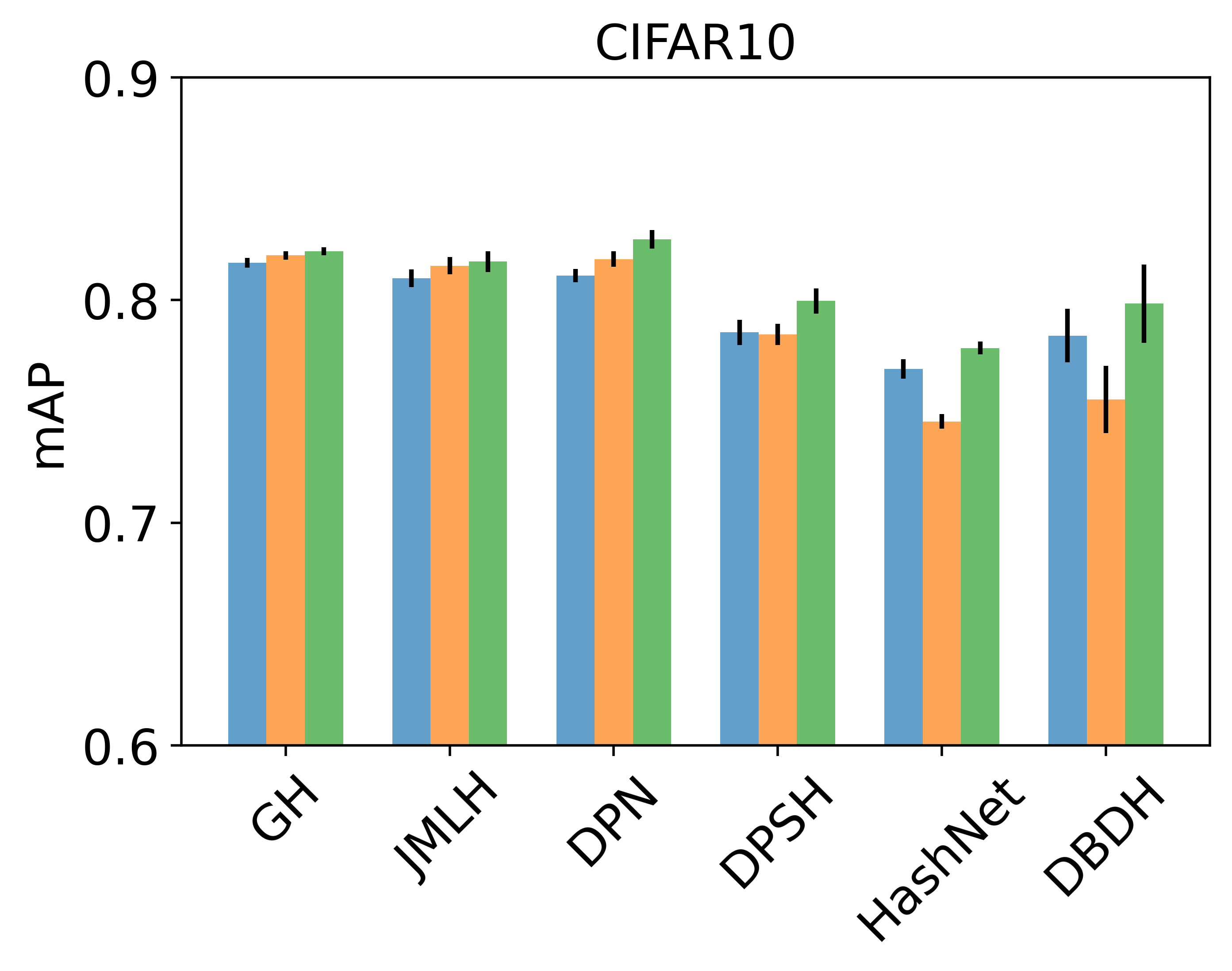}
		\caption{CIFAR10}
	\end{subfigure}
	\begin{subfigure}[b]{0.28\textwidth}
		\centering
		\includegraphics[width=\textwidth]{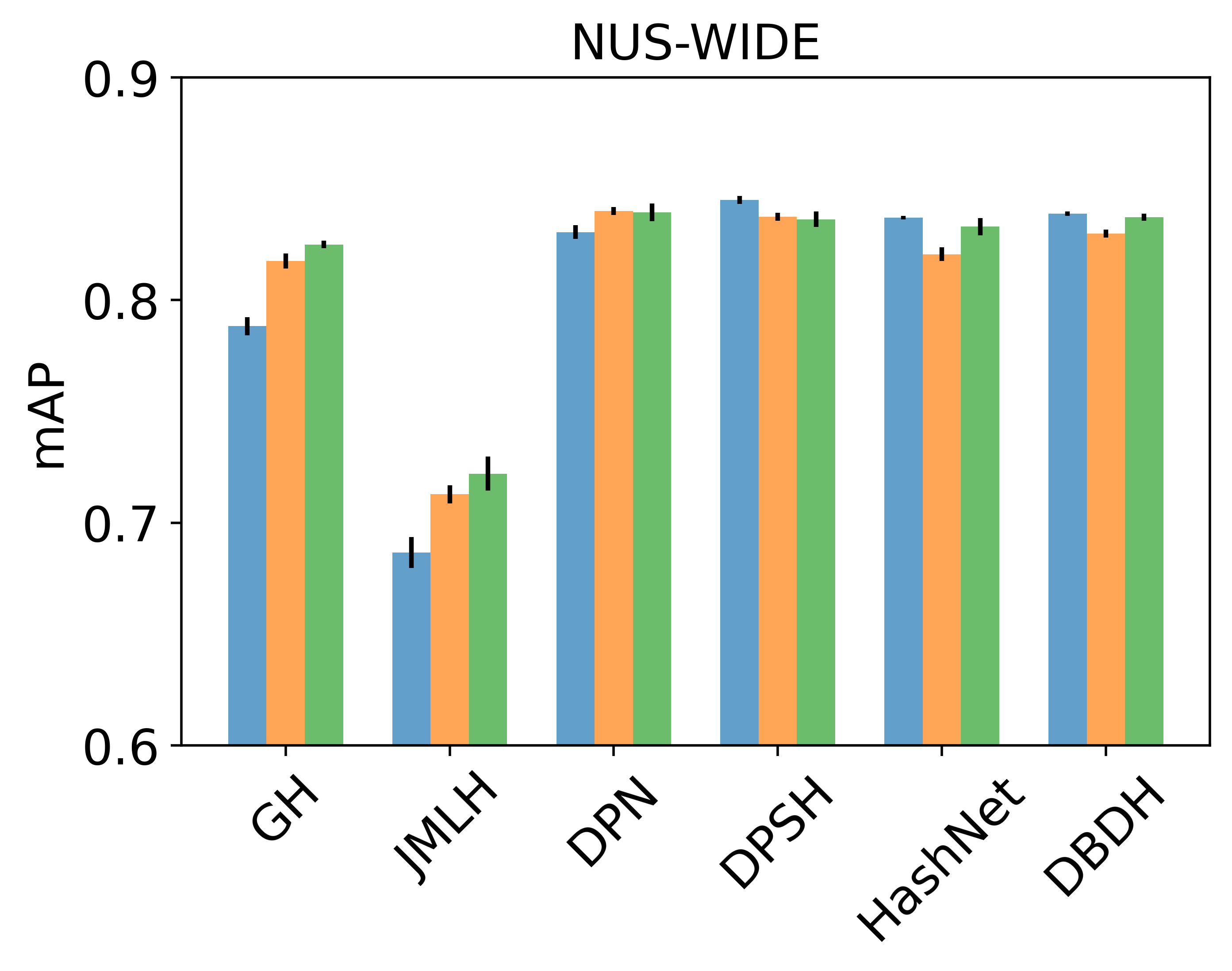}
		\caption{NUS-WIDE}
	\end{subfigure}
	\begin{subfigure}[b]{0.28\textwidth}
		\centering
		\includegraphics[width=\textwidth]{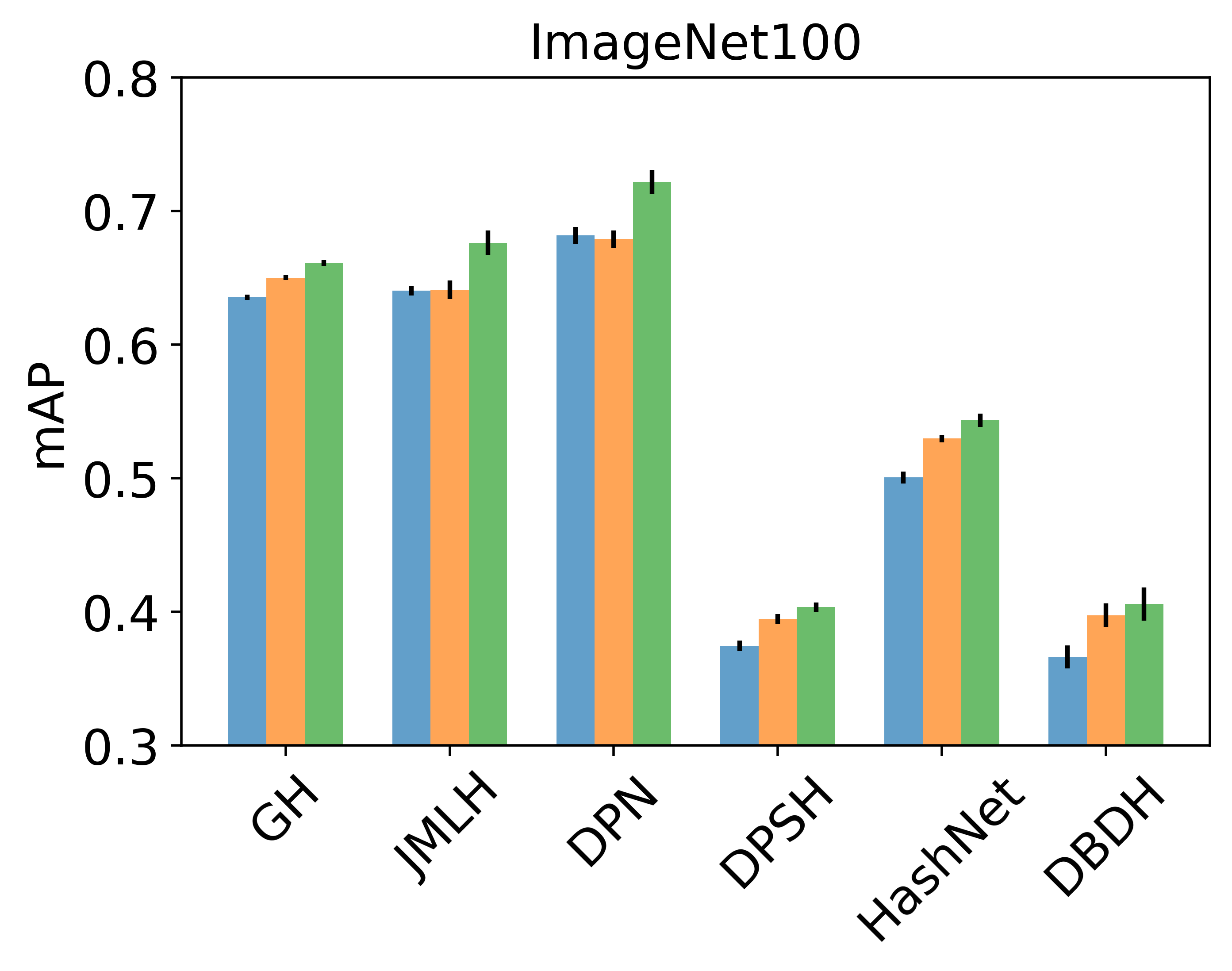}
		\caption{ImageNet100}
	\end{subfigure}
	\caption{Summary of mean average precision averaged over all number of bits for different dataset.}
	\label{fig:mapbar}
\end{figure*}

\begin{figure*}[ht]
	\centering
	\begin{subfigure}[b]{0.32\linewidth}
		\centering
		\includegraphics[width=\linewidth]{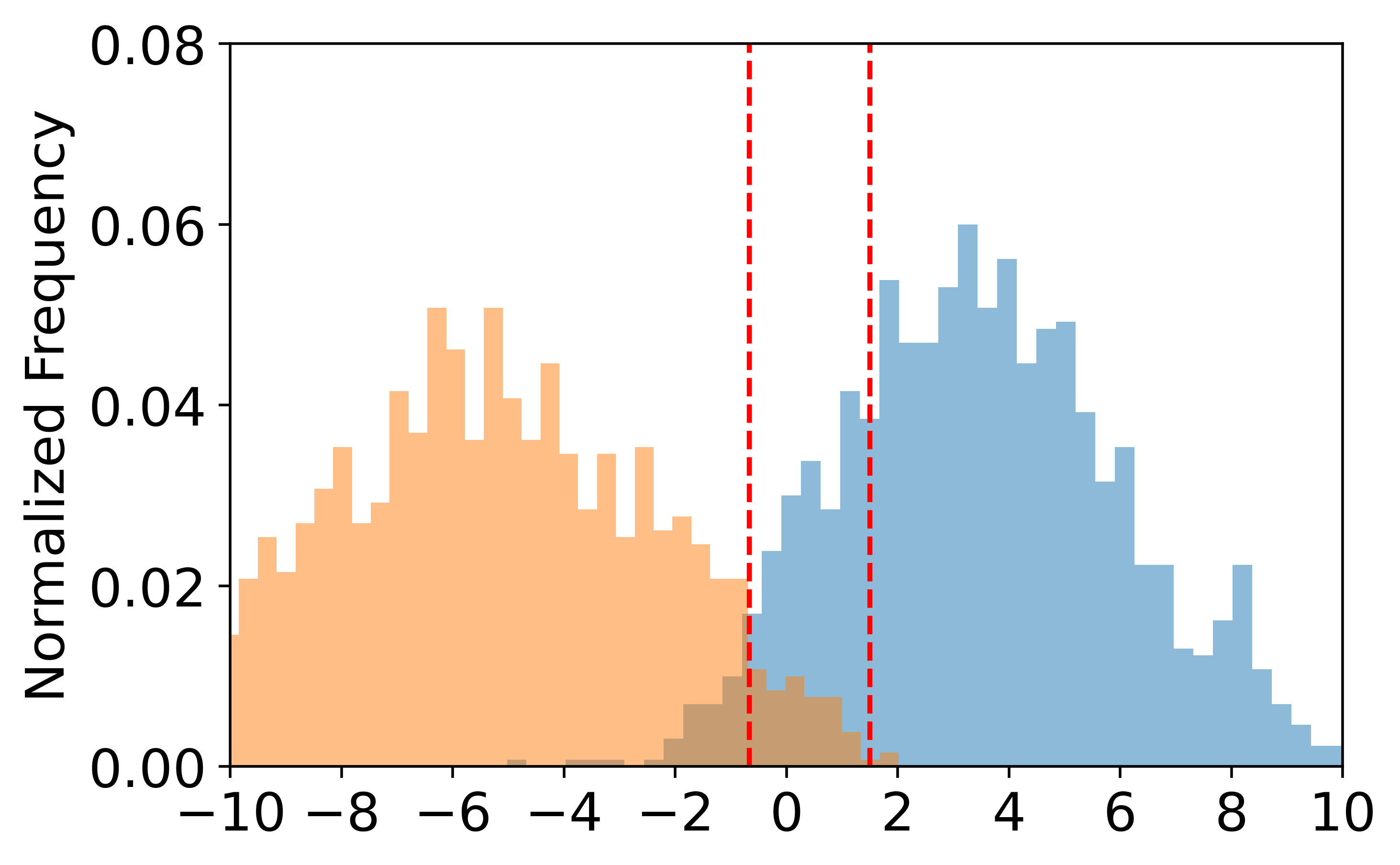}
		\caption{}
	\end{subfigure}
	\begin{subfigure}[b]{0.32\linewidth}
		\centering
		\includegraphics[width=\linewidth]{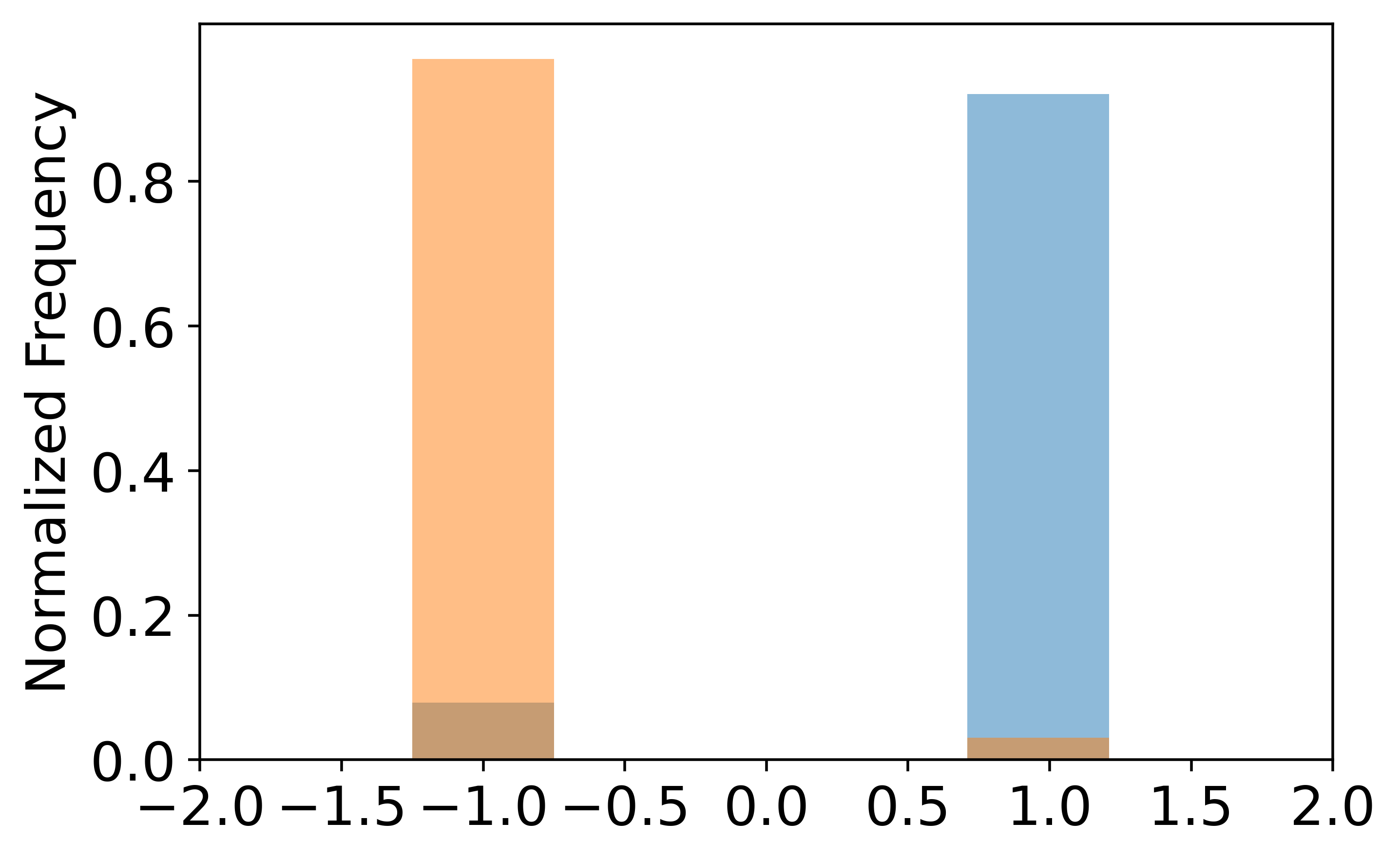}
		\caption{}
	\end{subfigure}
	\begin{subfigure}[b]{0.32\linewidth}
		\centering
		\includegraphics[width=\linewidth]{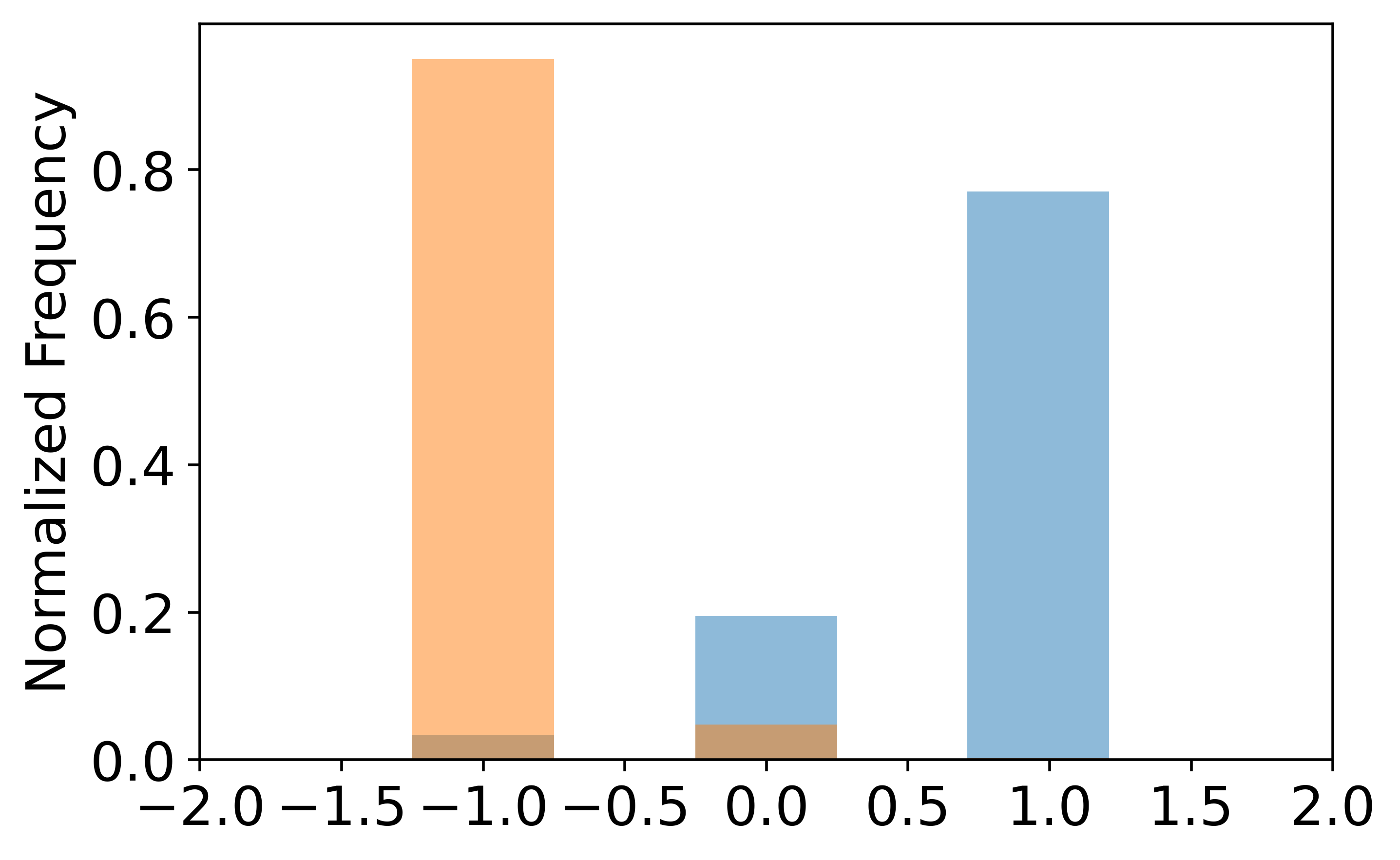}
		\caption{}
	\end{subfigure}
	\caption{Histograms of one of the bits of (a) raw output (b) binary hash codes and (c) ternary hash codes respectively. The vertical red dotted line is the double thresholds found by our proposed method in Algorithm \ref{alg:find_mu}. Orange and blue colors stand for two different classes. X-axis represented the range of the outputs.}
	\label{fig:visualize_dist}
\end{figure*}

\subsubsection{Datasets}
\textbf{CIFAR10.} CIFAR10 \cite{Krizhevsky09learningmultiple} consists of 60K 32x32 colored images in 10 classes. We follow the settings from \cite{HashNet_CaoICCV2017} and randomly select 100 images per class as the query set while the remaining images are used as the retrieval set. Besides, we randomly sample 500 images per class from the retrieval set as the training set.

\textbf{NUS-WIDE.} NUS-WIDE \cite{nus_wide_civr09} consists of 269K multi-labeled images in 81 concepts. We follow the settings from \cite{HashNet_CaoICCV2017} and use the most frequent 21 concepts as image annotations, thus 195K images are selected. We randomly select 100 images per concepts as the query set while the remaining images are used as the retrieval set. Moreover, we randomly sample 500 images per concepts from the retrieval set as the training set.

\textbf{ImageNet100.} ImageNet100 is a subset of ImageNet \cite{ILSVRC15} with only 100 classes. We follow the settings from \cite{HashNet_CaoICCV2017} and use the 100 classes. We use all the validation images from 100 classes as the query set and randomly sample 13K images from the retrieval set which consists of 128K images.

\subsubsection{Deep Hashing Methods}

For comparison, we have implemented six existing deep binary hashing methods for comparison as shown in Table \ref{tab:performance}\footnote{For some methods in the table, we are unable to achieve the performance with the hyperparameters setting from original paper, therefore we report the best performance that we can achieve.}. 
These binary hashing methods can be categorized into classification-based and pairwise-based methods. 
We have implemented Deep Polarized Network (DPN) \cite{fandeep}, GreedyHash (GH) \cite{GreedyHash_SuNIPS18}, and Just-Maximizing-Likelihood Hashing (JMLH) \cite{JMLH_ShenICCV19w} for classification-based methods while HashNet \cite{HashNet_CaoICCV2017}, Deep Balanced Discrete Hashing (DBDH) \cite{zheng2020deep} and Deep Pairwise-Supervised Hashing (DPSH) \cite{FDSH_LiIJCAI16} for pairwise-based methods.
Finally, we resized the input to 224x224 and we adopt AlexNet \cite{krizhevsky2017alexnet} as our backbone network, where image features are extracted from the \textit{fc7} layer. Then, the extracted features are act as input for the deep hashing methods and perform the respective training
\footnote{For detailed training hyperparameters of each deep hashing methods, please refer to the supplementary material.}. 

\subsection{Hashing Performance}\label{subsect:exper-perf}

To investigate the performance of image retrieval using ternary hashing, we use the standard top-k mean-Average Precision (mAP@k). We adopt the popular settings of $k=all,5000,1000$ for \textbf{CIFAR10}, \textbf{NUS-WIDE} and \textbf{ImageNet100} respectively according to \cite{zhu2016deep,fandeep,GreedyHash_SuNIPS18}. Main experimental results are summarized in Table \ref{tab:performance} and Figure \ref{fig:mapbar}, from which we made following observations: 

\begin{itemize}
	\item In majority of settings, ternary hashing based on Kleene logic demonstrated consistent improvements of hashing accuracies (in mAP) compared with six state-of-the-art binary hashing methods. The improvements are more pronounced for ImageNet100 than other datasets, for which the largest improvement is as high as 5.9\% (see bold numbers in Table \ref{tab:performance}). We ascribe the significant improvement to the reduced neighborhood ambiguity for challenging scenarios with large number of classes (100) and short bit lengths (16).  
		
	\item Ternary hashing based on Łukasiewicz logic was not as good as Kleene logic, and sometimes was outperformed by binary hashing.  Performance drops might be ascribed to the overconfidence and incurred errors in assigning distance 0 between two intermediate UNKNOWN states (see Table \ref{tab:lukas}).   It is interesting to further investigate influence of subtle difference between Łukasiewicz and Kleene logic for other hashing tasks. 
	
\end{itemize}
These experimental results justified our view that ternary hashing is a principled approach that can be adopted to improve any existing hashing method, although the extent of improvements might be dependent on the datasets, the length of hashing codes and the original hashing methods.

\subsection{Case Study}

\begin{figure}[h]
	\centering
	\begin{subfigure}[b]{\linewidth}
		\centering
		\fbox{
			\raisebox{0.3\height}{\includegraphics[width=0.1\linewidth]{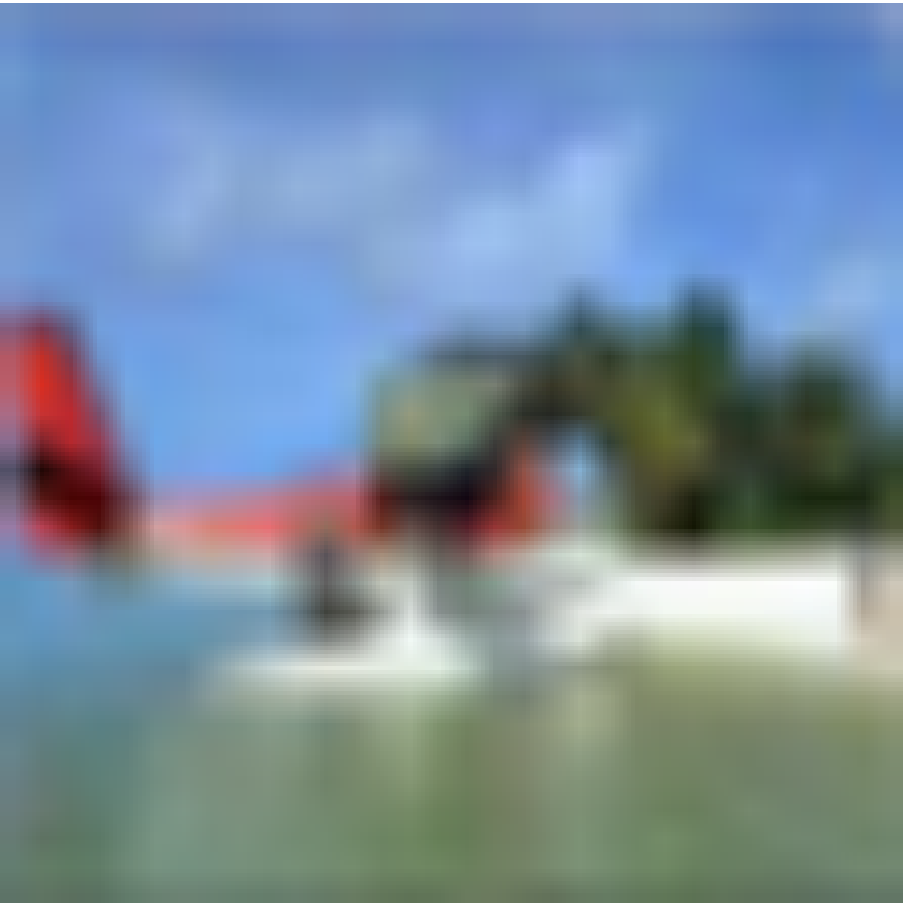}} 
			\hspace{0.2pt}
			\includegraphics[width=0.85\linewidth]{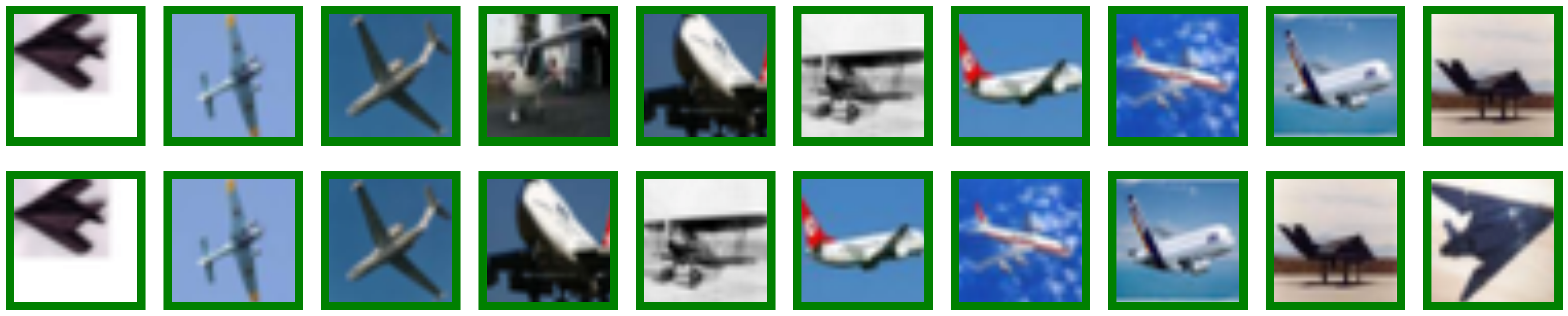}
		}
	\end{subfigure}
	\begin{subfigure}[b]{\linewidth}
		\centering
		\fbox{
			\raisebox{0.3\height}{\includegraphics[width=0.1\linewidth]{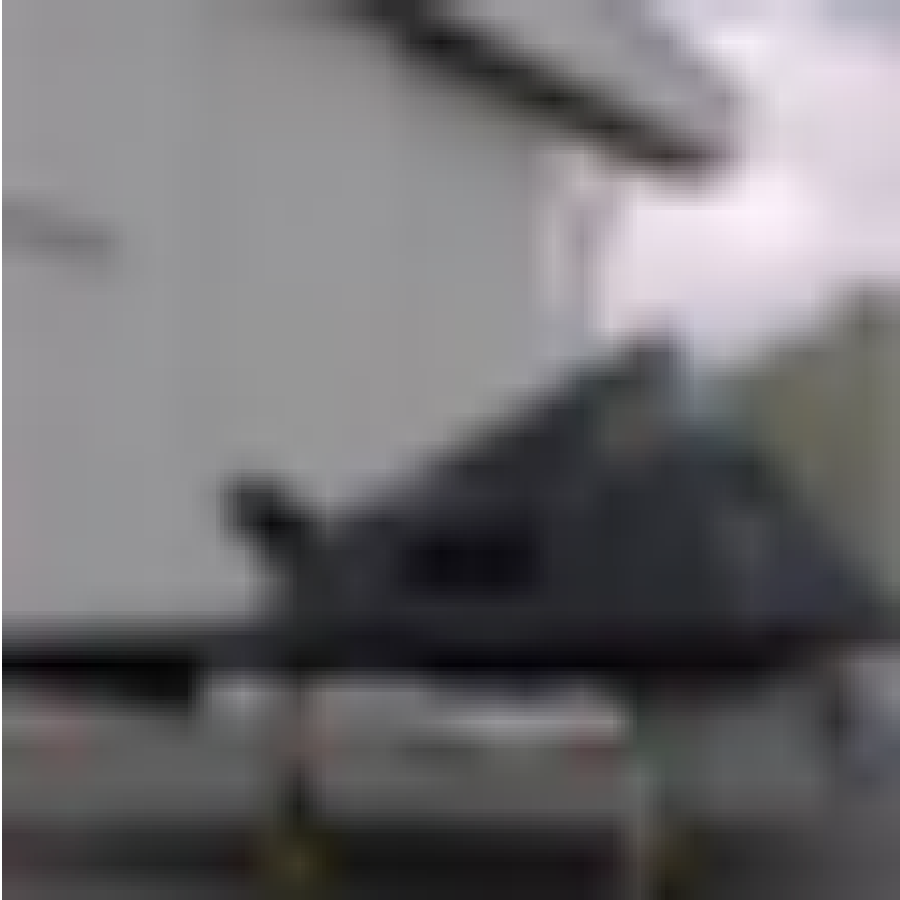}}
			\hspace{0.2pt}
			\includegraphics[width=0.85\linewidth]{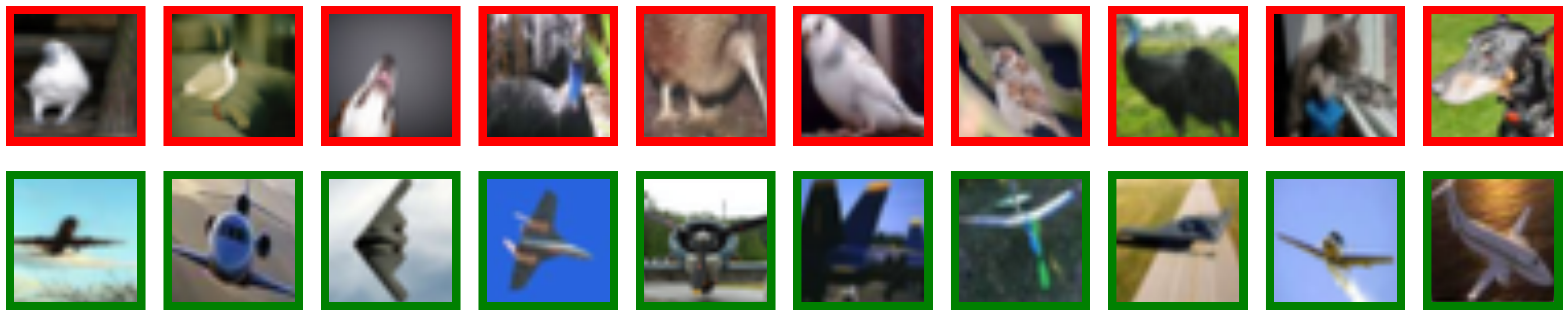}
		}
	\end{subfigure}
	\caption{Examples of top 10 retrieved images. For each query image (left), the top row represents retrieved images using binary hash codes while the bottom row represents the ternary hash codes. Green box stands for retrieved correctly while red box stands for retrieved wrongly. 
	}
	\label{fig:retrieval}
\end{figure}

\begin{figure}[h]
	\centering
	\includegraphics[width=0.4\linewidth]{legend.png}
	\\
	\begin{subfigure}[b]{0.48\linewidth}
		\centering
		\includegraphics[width=\linewidth]{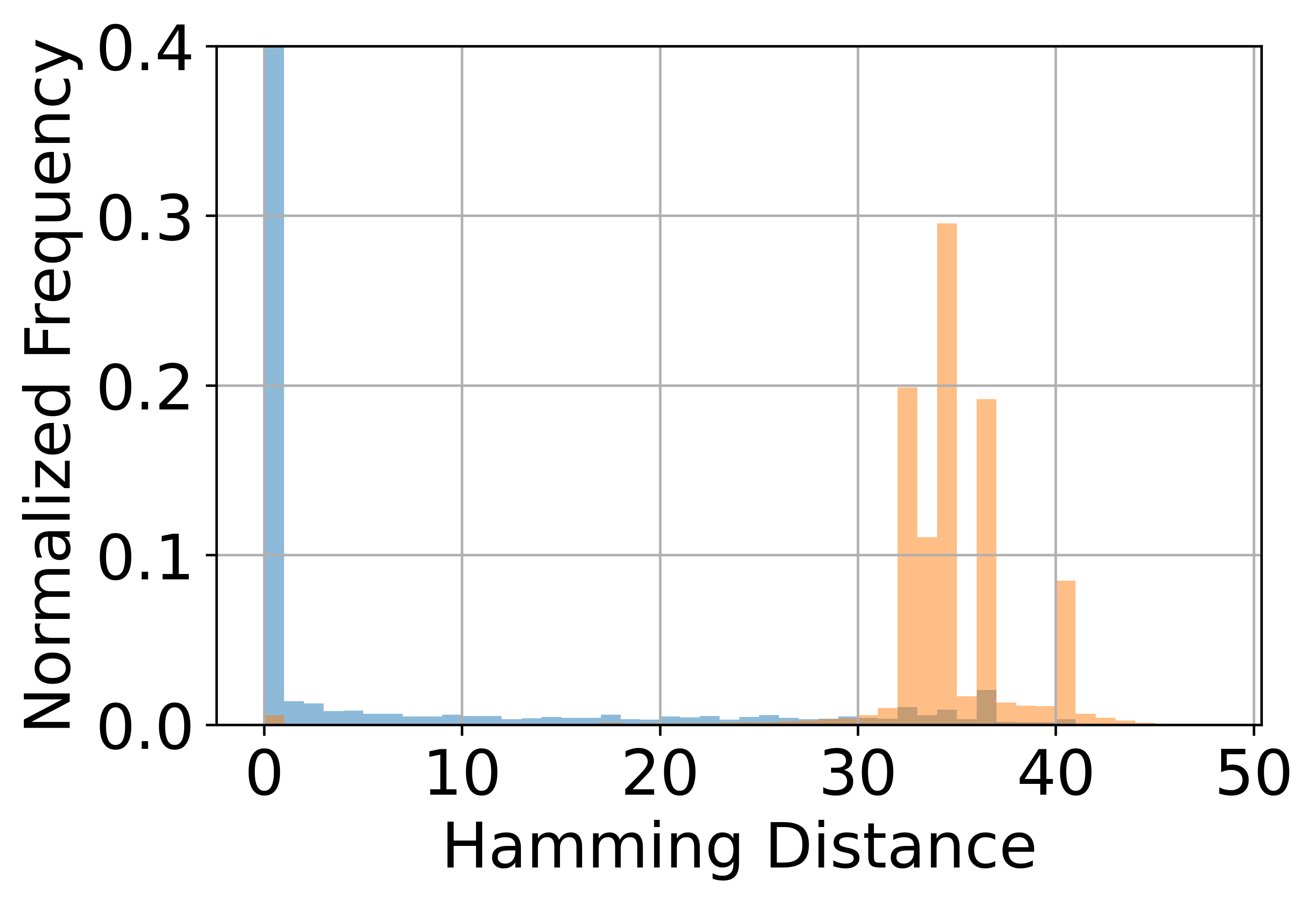}
		\caption{Binary Hash Codes}
		\label{fig:ambiguity_b}
	\end{subfigure}
	\begin{subfigure}[b]{0.48\linewidth}
		\centering
		\includegraphics[width=\linewidth]{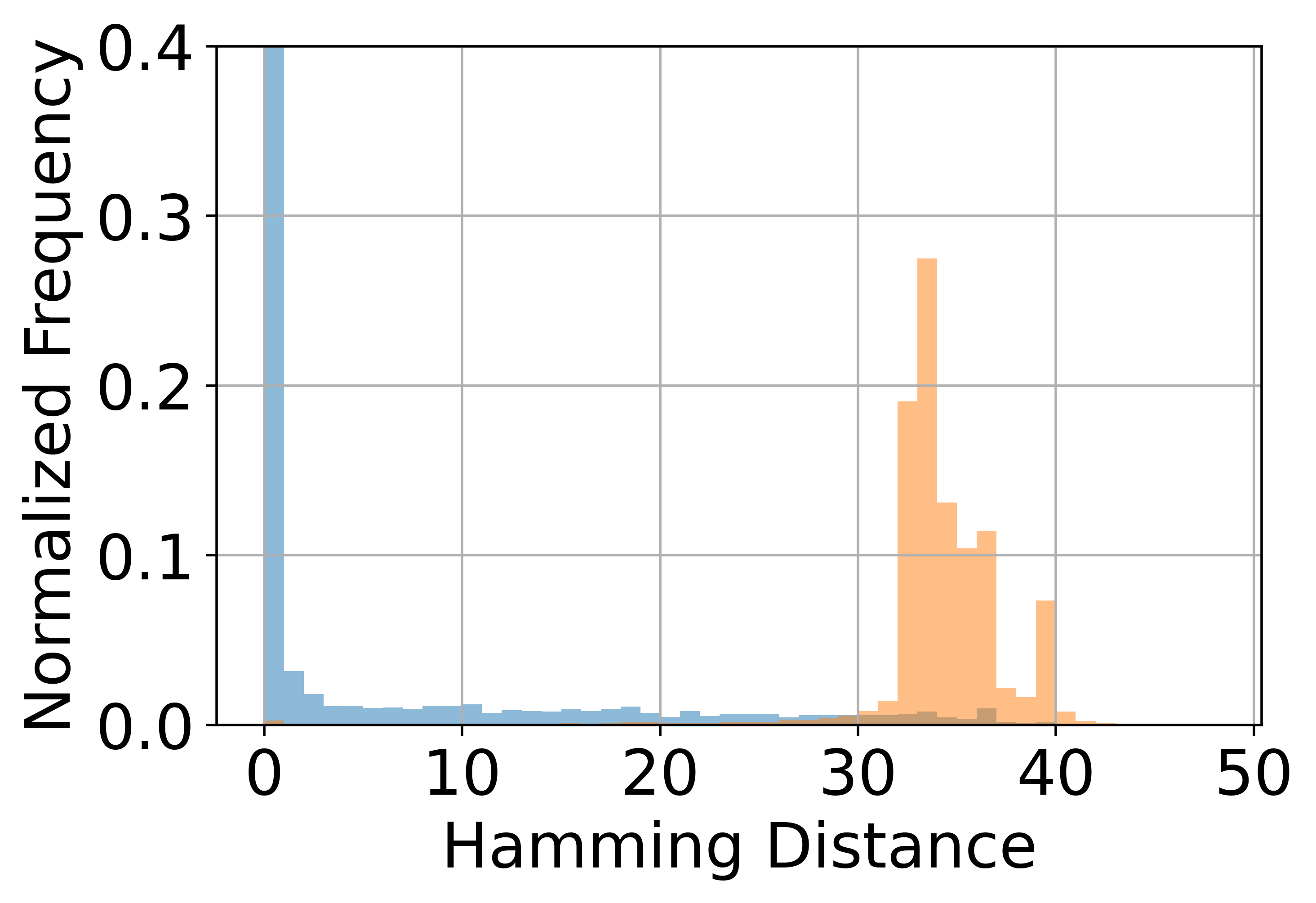}
		\caption{Ternary Hash Codes}
		\label{fig:ambiguity_t}
	\end{subfigure}
	\caption{The histogram of hamming distance between query hash codes and both positive and negative hash codes. The overlapping area of histogram of hamming distance was reduced from (a) 0.10 to (b) 0.08.
	\vspace{-1em}}
	\label{fig:ambiguity}
\end{figure}




Figure \ref{fig:retrieval} illustrates the top 10 CIFAR10 images retrieved for two query images using DPN-64 and DPN-64-K. It was shown that binary hashing retrieved wrong images due to mis-classifications, while ternary hashing is able to deal with challenging query images and successfully retrieve correct images. The reduced errors in challenging cases is also illustrated in Figure \ref{fig:visualize_dist}, which shows the errors due to binary code assignments is mitigated by introducing an intermediate region between the double-thresholds. Figure \ref{fig:ambiguity} shows that, for ternary hashing,  hamming distances between the query and negative samples increases (more concentrated at 32 in Figure \ref{fig:ambiguity_t}) while hamming distance with positive samples decreases (more concentrated to the left). Overlapping between histograms are thus reduced from 0.10 to 0.08, indicating a reduction in ambiguous region and a better separability between positive/negative samples. This observation is in accordance with  improved hashing accuracies illustrated in Figure \ref{fig:retrieval} and Table \ref{tab:performance}.

\subsection{Retrieval Efficiency on Binary Machines}

Retrieval efficiencies are measured by the time of calculating Hamming distance distance among samples. We use the double length of hash code for binary hashing to compare with the ternary ones, i.e. 16/32 denotes comparing 16-trit ternary hash code with 32-bit binary hash code, since ternary hash code was implemented by the double length binary code. The experiments are run on CPU in order to eliminate the impact of GPU acceleration. In this experiment, we count the calculation time of Hamming distance for 1 query sample to 10000 database samples $(1 @ 10000)$, and the unit is seconds. As Table \ref{efficiency} shows, with the proposed binary implementation, the efficiency of calculating ternary Hamming distance is comparable with calculating binary Hamming distance, and is much faster than matrix multiplication (matmul) using floating-point operation. Moreover, it was shown that the calculation of Kleene Hamming distance is slower than the Łukasiewicz Hamming distance, due to the extra step of calculating $t_b$ in eq. (\ref{KleeneEQ}).  

Finally, one must note that the measured efficiency is only applicable to ternary hashing implementation on binary machines.  Possible improvements in efficiency brought by dedicated hardware supporting ternary logic is discussed in Sect. \ref{sect-dis-conl}. 

\begin{table}[t]
	\centering
	\resizebox{0.9\linewidth}{!}{
		\begin{tabular}{|c|c|c|c|c|c|}
			\hline
			\multicolumn{6}{|c|}{1 @ 10000}                       \\ \hline
			& \textbf{16/32}  & \textbf{32/64}  & \textbf{64/128} & \textbf{128/256} & \textbf{256/512} \\ \hline
			matmul & 2.05 & 3.62 & 8.40 & 15.61 & 28.48 \\ \hline
			Kleene & 0.18 & 0.17 & 0.19 & 0.27  & 0.36  \\ \hline
			Łukasiewicz   & 0.11 & 0.12 & 0.13 & 0.14  & 0.18  \\ \hline
			\hline
			binary & 0.10 & 0.10 & 0.13 & 0.16  & 0.12  \\ \hline
	\end{tabular}}
	\caption{Retrieval Time measured in second.}
	\label{efficiency}
\end{table}

\section{Discussion and Conclusion}\label{sect-dis-conl}
\label{discussion}

\begin{figure}[t]
    \centering
    \includegraphics[width=0.7\linewidth]{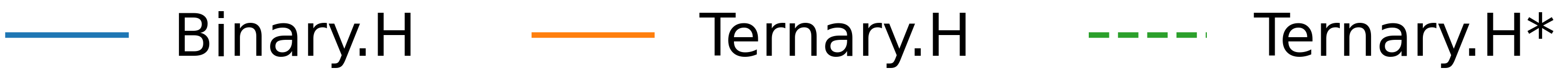} \\
    \includegraphics[width=0.6\linewidth]{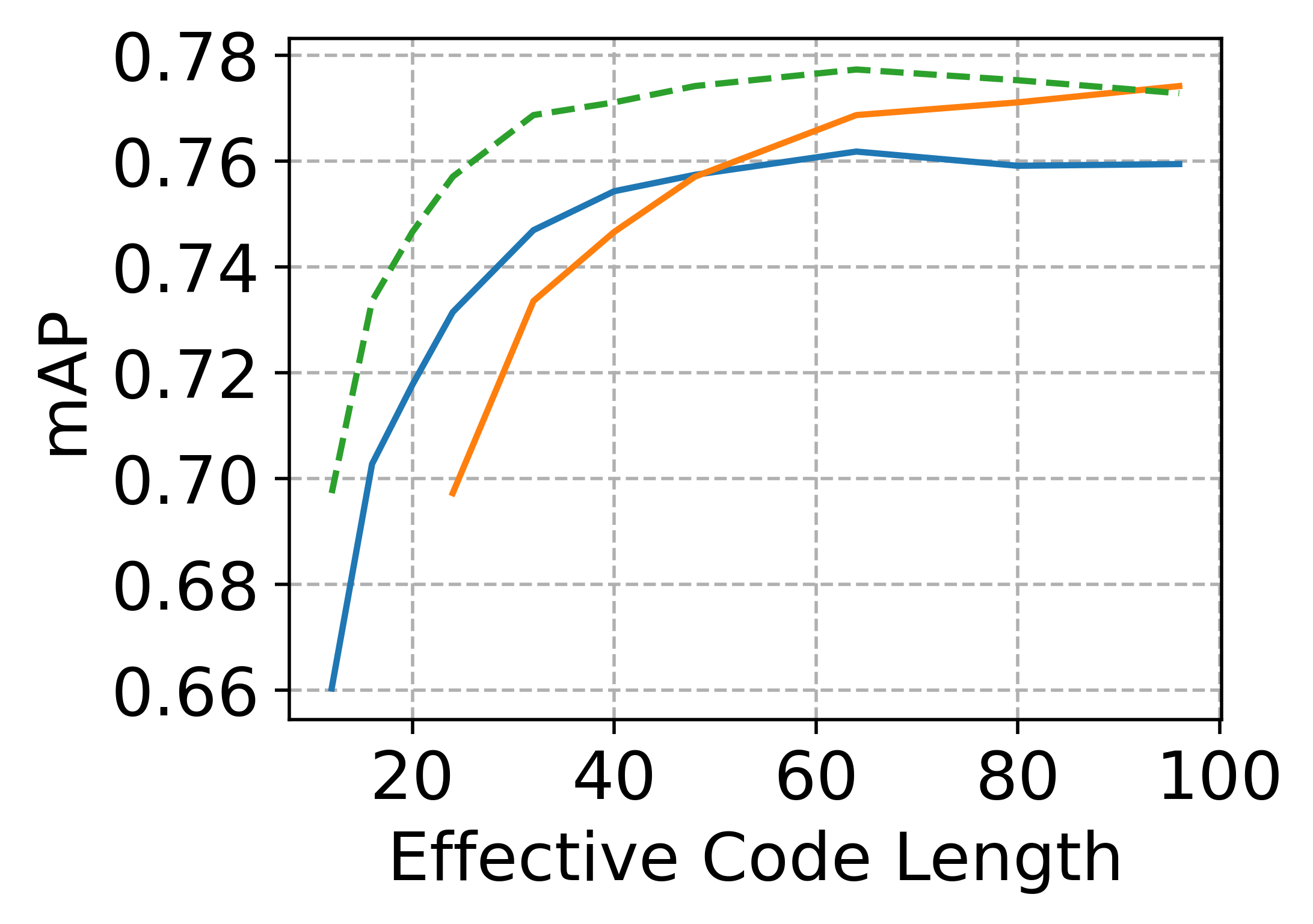}
    \caption{A summary of retrieval performance (mAP) averaged among different datasets and methods over different number of code length. \textbf{Binary.H} and \textbf{Ternary.H} means Binary and Ternary Hash Codes respectively. \textbf{*} denotes the performance if ternary hashing is implemented on a genuine ternary machine. Effective code length represents the number of bit/trits on a binary machine or a ternary machine.}
    \label{fig:linechart}
\end{figure}


This paper demonstrates a novel ternary hashing method, which aims to improve existing deep binary hashing models by introducing an intermediate UNKNOWN state to reduce errors ascribed to ambiguous binary code assignments.  The reduced errors in each bit assignment lead to better separability between Hamming distance histograms for positive and negative data points, and consequently, hashing accuracies are consistently improved ranging from 1\% to 5.9\%.  


In this paper an implementation of ternary hashing on binary machines is provided to demonstrate that ternary hashing can take advantages of widely available binary systems. Since the implementation is based on double-bits binary code and omits '11' in encoding, there is an inevitable 25\% loss of coding efficiency. Nevertheless, the performance of 24-trit ternary hashing with Kleene logic (75\% coding efficiency) has consistently exceeded the 32-bit binary hashing as shown in Table \ref{tab:performance}. If ternary hashing is implemented on genuine ternary machines and exploit 100\% coding efficiency (the green dotted line in Figure \ref{fig:linechart}), we assume that the performance of ternary hashing 
would also comprehensively surpass that of binary hashing. There are indeed dedicated implementations of ternary logic with Field Programmable Gate Array (FPGA) in \cite{stephen_fpga_92} \cite{boucle_fpl_fpga_17}. As future work, it is interesting to explore and adopt such an implementation to achieve full potential of ternary hashing.


{
\bibliographystyle{ieee_fullname}
\bibliography{Learn2Hash}
}

\end{document}


\title{Supplementary Material \\ Ternary Hashing}


\author{Kam Woh Ng\\
WeBank AI\\
Shenzhen\\
{\tt\small jinhewu@webank.com}
\and
Chang Liu\\
Xi'an Jiaotong-Liverpool University\\
Suzhou\\
{\tt\small chang.liu17@student.xjtlu.edu.cn}
\and
Lixin Fan\\
WeBank AI\\
Shenzhen\\
{\tt\small lixinfan@webank.com}
\and
Yilun Jin\\
The Hong Kong University of Science and Technology\\
Hong Kong\\
{\tt\small yilun.jin@connect.ust.hk}
\and
Ce Ju\\
WeBank AI\\
Shenzhen\\
{\tt\small ceju@webank.com}
\and
Tianyu Zhang\\
WeBank AI\\
Shenzhen\\
{\tt\small brutuszhang@webank.com}
\and
Chee Seng Chan\\
University of Malaya\\
Kuala Lumpur\\
{\tt\small cs.chan@um.edu.my}
\and
Qiang Yang\\
The Hong Kong University of Science and Technology\\
Hong Kong\\
{\tt\small qyang@cse.ust.hk}
}

\maketitle

\section{Introduction}

Appendix \ref{sec:hyperparam} describes the implementation and hyperparameters of various deep hashing methods mentioned in the main paper. Appendix \ref{sec:method} is the detailed explanation and visualization of our proposed algorithm.

\section{Implementation and Hyperparameters}
\label{sec:hyperparam}

For comparison, we have implemented six existing deep binary hashing methods for comparison and they can be categorized into classification-based and pairwise-based methods. The methods include Deep Polarized Network (DPN)\footnote{\url{https://github.com/swuxyj/DeepHash-pytorch/blob/master/DPN.py}} \cite{fandeep}, GreedyHash (GH)\footnote{\url{https://github.com/ssppp/GreedyHash}} \cite{GreedyHash_SuNIPS18}, and Just-Maximizing-Likelihood Hashing (JMLH)\footnote{\url{https://github.com/ymcidence/TBH/blob/master/model/jmlh.py}} \cite{JMLH_ShenICCV19w} for classification-based methods while HashNet\footnote{\url{https://github.com/thuml/HashNet/tree/master/pytorch}} \cite{HashNet_CaoICCV2017}, Deep Balanced Discrete Hashing (DBDH)\footnote{\url{https://github.com/swuxyj/DeepHash-pytorch/blob/master/DBDH.py}} \cite{zheng2020deep} and Deep Pairwise-Supervised Hashing (DPSH)\footnote{\url{https://github.com/jiangqy/DPSH-pytorch}} \cite{FDSH_LiIJCAI16} for pairwise-based methods. The footnote indicates the github repository that we also refered to during the implementation. However, we couldn't reproduce the same result for some methods even though using the setting in original paper. Therefore we performed a grid-search for the hyperparameters and the best hyperparameters we found are summarized in Table \ref{tab:hyperparameter}.

For all the implementations, we have used Python 3.7 and PyTorch \cite{pytorch} as our deep learning framework. For dataset, we followed previous papers \cite{zhu2016deep,fandeep,GreedyHash_SuNIPS18,HashNet_CaoICCV2017,JMLH_ShenICCV19w} which used CIFAR-10, NUS-WIDE and ImageNet100 as their benchmark, the dataset can also be found here\footnote{\url{https://github.com/thuml/HashNet/tree/master/pytorch/data}}.

\begin{table*}
	\centering
	\begin{tabular}{c|c|c|c|c|c}
		\toprule
		Method & \centering Method Hyperparameter & Optimizer & LR & Batch Size & Epochs \\
		\midrule
		
		GreedyHash & $\alpha_{imgnet, nuswide} = 1$; $\alpha_{cifar10} = 0.1$ ; $p=3$ & SGD & 0.001 & 256 & 200 \\ 
		\midrule
		JMLH & $\lambda_{imgnet} = 0.01$ ; $\lambda_{nuswide}=0.005$ ; $\lambda_{cifar10} = 0.1$ & Adam & 0.0001 & 256 & 200 \\ 
		\midrule
		DPN & $m = 1$ & SGD & 0.0001 & 256 & 200 \\ 
		\midrule
		DPSH & $\eta = 0.1$ & SGD & 0.001 & 64 & 200 \\ 
		\midrule
		HashNet & $\alpha = 1$ & SGD & 0.002 & 256 & 200 \\ 
		\midrule
		DBDH & $\alpha=0.1$; $p=2$ & SGD & 0.001 & 64 & 200 \\ 
		\bottomrule
	\end{tabular}
	\caption{The hyperparameter used for training the network with the binary hashing methods. }
	\label{tab:hyperparameter}
\end{table*}



\section{Ternary Hamming Distance}
\label{sec:method}

\subsection{Expectation of Ternary Hamming Distance}
\label{sec:thd}

\begin{table}
	\centering
	\begin{minipage}{.45\linewidth}
		\centering
		\resizebox{\linewidth}{!}{%
			\begin{tabular}{|c|c|ccc|c|}
				\hline
				\multicolumn{2}{|c|}{}  & \multicolumn{3}{c|}{$a \leftrightarrow b$} & $\neg a$ \\ 
				\hline
				\multicolumn{2}{|c|}{$b$} & \textbf{+1} & \textbf{0} & \textbf{-1} & \\ 
				\hline
				\multirow{3}{*}{$a$} & \textbf{+1} & +1 & 0 & -1 & -1 \\ 
				& \textbf{0} & 0 & +1 & 0 & 0 \\ 
				& \textbf{-1} & -1 & 0 & +1 & +1 \\ 
				\hline
			\end{tabular}%
		}
		\subcaption{Luka logic}
	\end{minipage}%
	\hfill
	\begin{minipage}{.45\linewidth}
		\centering
		\resizebox{\linewidth}{!}{%
			\begin{tabular}{|c|c|ccc|c|}
				\hline
				\multicolumn{2}{|c|}{}  & \multicolumn{3}{c|}{$a \leftrightarrow b$} & $\neg a$ \\ 
				\hline
				\multicolumn{2}{|c|}{$b$} & \textbf{+1} & \textbf{0} & \textbf{-1} & \\ 
				\hline
				\multirow{3}{*}{$a$} & \textbf{+1} & +1 & 0 & -1 & -1 \\ 
				& \textbf{0} & 0 & 0 & 0 & 0 \\ 
				& \textbf{-1} & -1 & 0 & +1 & +1 \\ 
				\hline
			\end{tabular}%
		}
		\subcaption{Kleene logic}
	\end{minipage}
	
	\caption{The truth tables for ternary logic in terms of $+1, 0, -1$.}
	\label{tab:logictable}
\end{table}

\begin{table}
	\centering
	\begin{minipage}{.45\linewidth}
		\centering
		\resizebox{\linewidth}{!}{%
			\begin{tabular}{|c|c|ccc|}
				\hline 
				\multicolumn{2}{|c|}{$b$} & \rule[1em]{0pt}{0pt} \textbf{$b^T$} & \textbf{$b^U$} & \textbf{$b^F$} \\ 
				\hline
				\multirow{3}{*}{$a$} & \rule[1em]{0pt}{0pt} \textbf{$a^T$} & 0 & 0.5 & 1 \\ 
				& \rule[1em]{0pt}{0pt} \textbf{$a^U$} & 0.5 & 0 & 0.5 \\ 
				& \rule[1em]{0pt}{0pt} \textbf{$a^F$} & 1 & 0.5 & 0 \\ 
				\hline
			\end{tabular}%
		}
		\subcaption{Luka logic}
	\end{minipage}%
	\hfill
	\begin{minipage}{.45\linewidth}
		\centering
		\resizebox{\linewidth}{!}{%
			\begin{tabular}{|c|c|ccc|}
				\hline 
				\multicolumn{2}{|c|}{$b$} & \rule[1em]{0pt}{0pt} \textbf{$b^T$} & \textbf{$b^U$} & \textbf{$b^F$} \\ 
				\hline
				\multirow{3}{*}{$a$} & \rule[1em]{0pt}{0pt} \textbf{$a^T$} & 0 & 0.5 & 1 \\ 
				& \rule[1em]{0pt}{0pt} \textbf{$a^U$} & 0.5 & 0.5 & 0.5 \\ 
				& \rule[1em]{0pt}{0pt} \textbf{$a^F$} & 1 & 0.5 & 0 \\ 
				\hline
			\end{tabular}%
		}
		\subcaption{Kleene logic}
	\end{minipage}

	\caption{The elements in the table are the ternary hamming distance between trits $a$ and $b$. $a^T,a^U,a^F$ are $+1,0,-1$ respectively (similar vein for trits $b$).}
	\label{tab:hdtable}
\end{table}

For two ternary codes or \textit{trits} $a,b \in \{-1, 0, +1\}$, we propose to adopt \textit{ternary Hamming distance} $\mathsf{THD}(a, b)$ as follows,
\begin{equation}\label{ternaryHamm}
\mathsf{THD}(a, b) := \frac{1}{2}( \neg(a \leftrightarrow b) + 1 ) \rightarrow \{0, 0.5, 1\}, 
\end{equation}
in which operators $\neg, \leftrightarrow$ are given in Table \ref{tab:logictable}. While Table \ref{tab:hdtable} shows the ternary hamming distance between trits $a$ and $b$ given different assignments of ternary hash codes . 

For a set of samples, let $P^T_a, P^U_a, P^F_a$ denote the probabilities being assigned codes $+1, 0, -1$ respectively (and  $P^T_b, P^U_b, P^F_b$ being defined in a similar vein for another set of samples). According to the ternary hamming distance between trits $a$ and $b$ in Table \ref{tab:hdtable} and the assumption that assigned codes for trits $a$ and $b$ are independent, the expectation of ternary hamming distances is then given by, 

\begin{equation}
\begin{split}
\mathbb{E}_{L}(A, B) = & 1 \cdot \Big(P_a^T P_b^F + P_a^F P_b^T \Big)  + \\ 
& 0.5 \cdot \Big( P_a^U (P_b^T + P_b^F) +  P_b^U (P_a^T + P_a^F) \Big) + \\
& 0 \cdot \Big(P_a^T P_b^T + P_a^F P_b^F + P_a^U P_b^U \Big) \\
= & 1 \cdot \Big(P_a^T P_b^F + P_a^F P_b^T \Big)  + \\ 
& 0.5 \cdot \Big( P_a^U (P_b^T + P_b^F) +  P_b^U (P_a^T + P_a^F) \Big),
\label{eq:lukahd}
\end{split}
\end{equation}

\begin{equation}
\begin{split}
\mathbb{E}_{K}(A,B) = & 1 \cdot \Big(P_a^T P_b^F + P_a^F P_b^T \Big)  +  0.5 \cdot P_a^U P_b^U + \\ 
& 0.5 \cdot \Big( P_a^U (P_b^T + P_b^F) +  P_b^U (P_a^T + P_a^F) \Big) + \\
& 0 \cdot \Big(P_a^T P_b^T + P_a^F P_b^F\Big) \\
= & 1 \cdot \Big(P_a^T P_b^F + P_a^F P_b^T \Big)  +  0.5 \cdot P_a^U P_b^U + \\ 
& 0.5 \cdot \Big( P_a^U (P_b^T + P_b^F) +  P_b^U (P_a^T + P_a^F) \Big),
\label{eq:kleenhd}
\end{split}
\end{equation}


 

\subsection{The Objective Function}

\begin{table}[h]
	\centering
	\begin{tabular}{|c|c|c|c|c|}
		\hline 
		& $X_1$ & $X_2$ & $X_3$ & $X_4$ \\
		\hline
		$X_1$ & -1 & 1 & 1 & 1 \\
		\hline
		$X_2$ & 1 & -1 & 1 & 1 \\
		\hline
		$X_3$ & 1 & 1 & -1 & 1 \\
		\hline
		$X_4$ & 1 & 1 & 1 & -1 \\
		\hline
	\end{tabular}%
	\caption{This is the example of $(-1)^s$ given 4 different classes $X_1$, $X_2$, $X_3$ and $X_4$.}
	\label{tab:s}
\end{table}

Ternary hamming distances in (\ref{eq:lukahd}) and (\ref{eq:kleenhd}) are used to quantify (dis-)similarity between samples from two different classes. For samples belonging to $C$ different classes  $ X := X_1 \cup  \cdots \cup X_C$, the  pairwise ternary hamming distances over all possible pairs of sets is given by 
\begin{equation}\label{eq:pairTHD}
	\mathbb{E}_{K,L}(X) =  \sum_{A, B \in \{X_1, \cdots, X_C\}} (-1)^s \mathbb{E}_{K,L} (A, B), 
\end{equation}
where $s=\Big \{ \begin{array}{lr}
	0  & \text{if } A \neq B\\
	1 &  \text{if } A=B
\end{array}$ and the example is shown in Table \ref{tab:s}. Note that $\mathbb{E}_{K,L}(X)$ are respective Kleene and  Łukasiewicz ternary hamming distances, and we may omit the subscript for brevity. 

Using (\ref{eq:pairTHD}) as our objective function, the proposed double-threshold searching algorithm is to find, for each ternary hashing mapping, a pair of thresholds that \emph{maximize the expectation of pairwise ternary hamming distances between negative samples, minus ternary hamming distances between positive samples}. In other words, the maximization part is to add up the inter-class (e.g. $X_1 \text{ \& } X_2$) ternary hamming distances and the minimization part is to minus the intra-class (e.g. $X_1 \text{ \& } X_1$) ternary hamming distances.


\subsection{Reduced Ambiguity with Ternary Hashing}

\begin{figure}[h]
	\centering
	\begin{subfigure}[b]{0.45\linewidth}
		\centering
		\includegraphics[width=\linewidth]{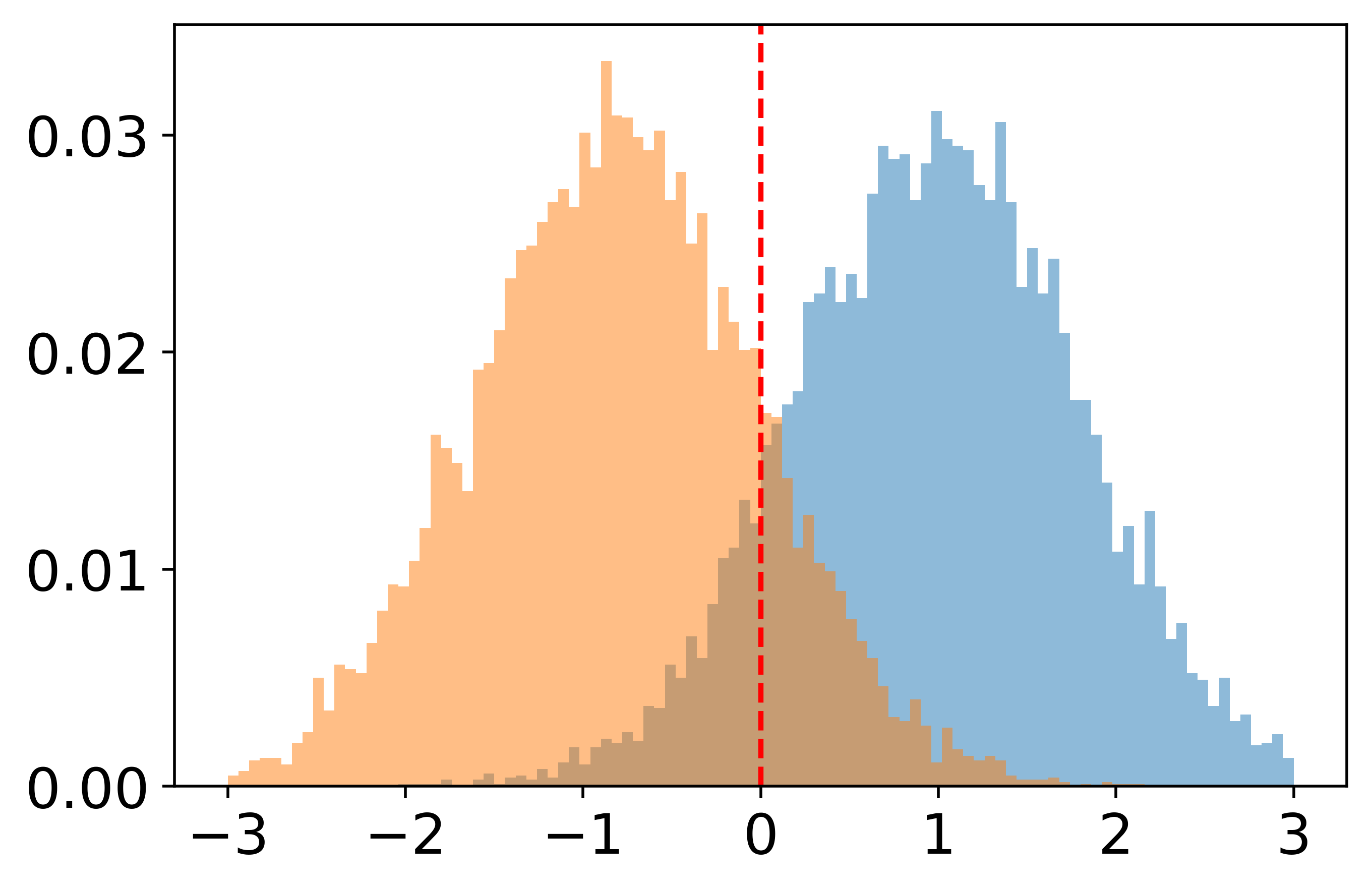}
	\end{subfigure}
	\begin{subfigure}[b]{0.45\linewidth}
		\centering
		\includegraphics[width=\linewidth]{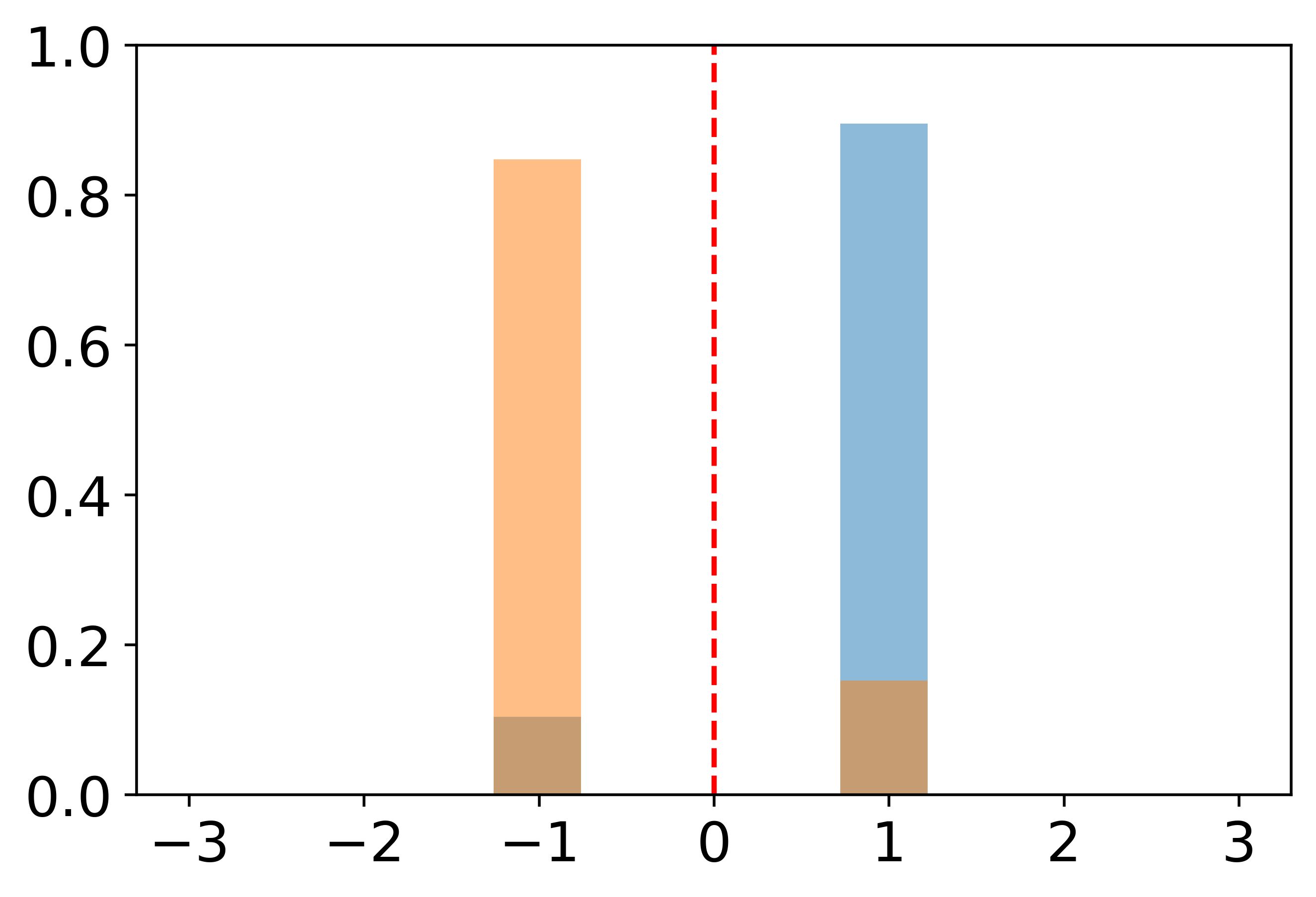}
	\end{subfigure}
	\caption{(Left) the histogram of the raw output of two classes $A$ and $B$. (Right) the histogram of the binary hash codes of two classes. The red dotted line is the threshold for mapping the raw output to the binary codes. The region at the left of the threshold is $P_a^F$ and $P_b^F$ and the region at the right is $P_a^T$ and $P_b^T$. X-axis is the range of outputs and Y-axis is the probability of each bin in the histogram (normalized from the frequencies).}
	\label{fig:bintwo}
\end{figure}

\begin{figure}[h]
	\centering
	\begin{subfigure}[b]{0.45\linewidth}
		\centering
		\includegraphics[width=\linewidth]{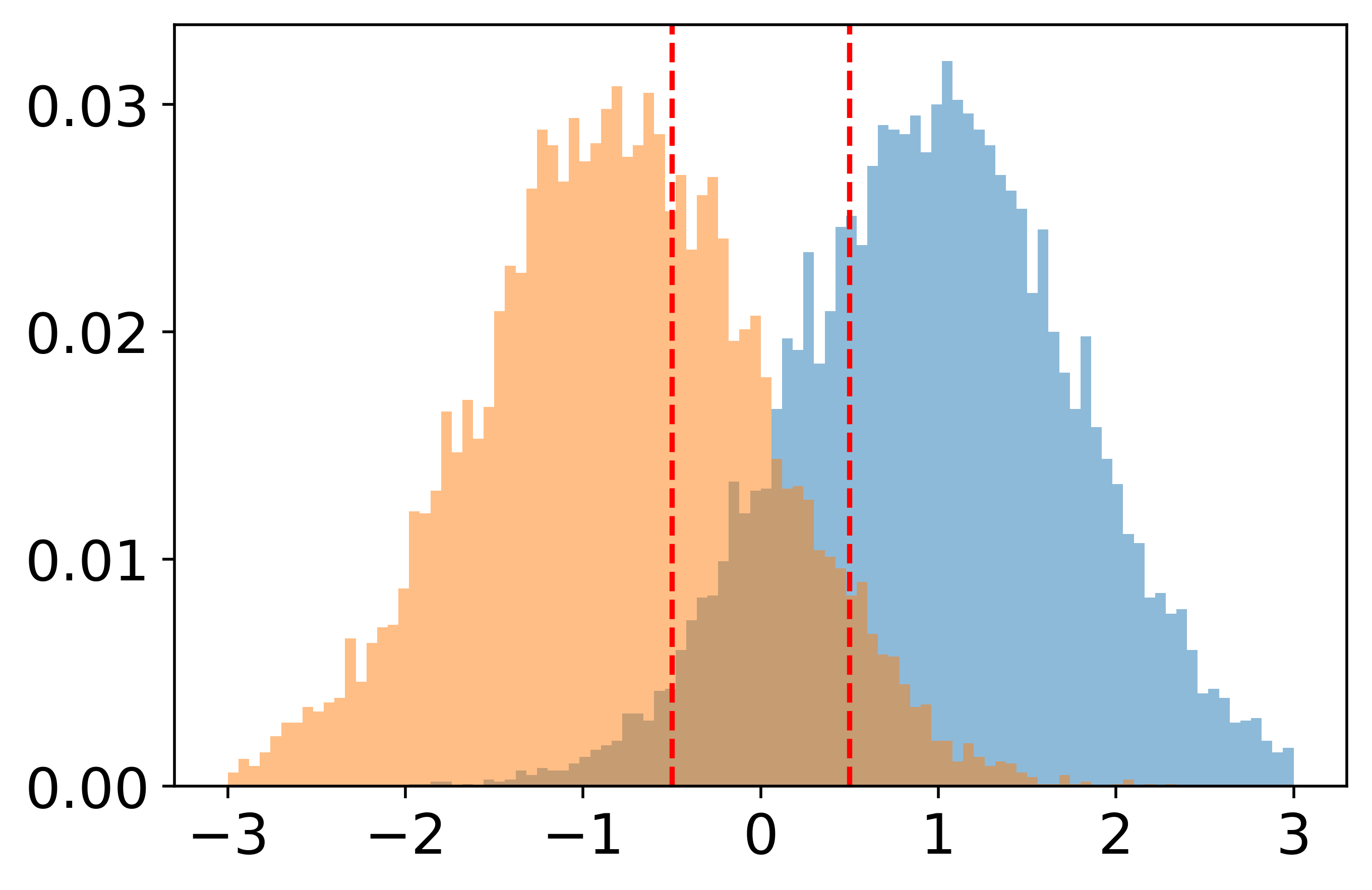}
	\end{subfigure}
	\begin{subfigure}[b]{0.45\linewidth}
		\centering
		\includegraphics[width=\linewidth]{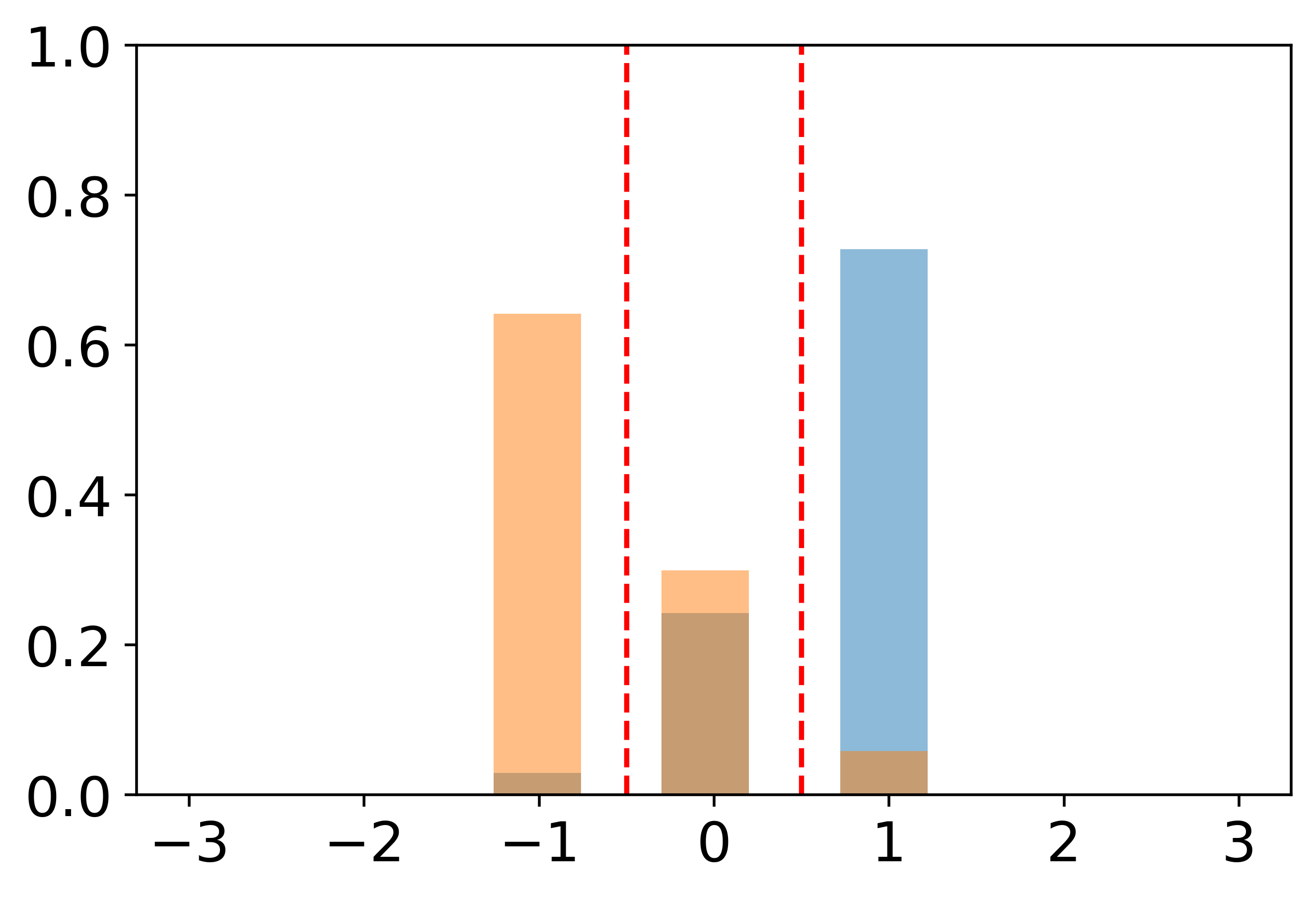}
	\end{subfigure}
	\caption{(Left) the histogram of the raw output of two classes $A$ and $B$. (Right) the histogram of the ternary hash codes of two classes. The two red dotted lines are the thresholds for mapping the raw output to the ternary codes. The region at the left of the threshold is $P_a^T$ and $P_b^T$, the region at the middle is $P_a^U$ and $P_b^U$ and the region at the right is $P_a^F$ and $P_b^F$. X-axis is the range of outputs and Y-axis is the probability of each bin in the histogram (normalized from the frequencies).}
	\label{fig:tertwo}
\end{figure}

\begin{figure}[h]
	\centering
	\begin{subfigure}[b]{0.45\linewidth}
		\centering
		\includegraphics[width=\linewidth]{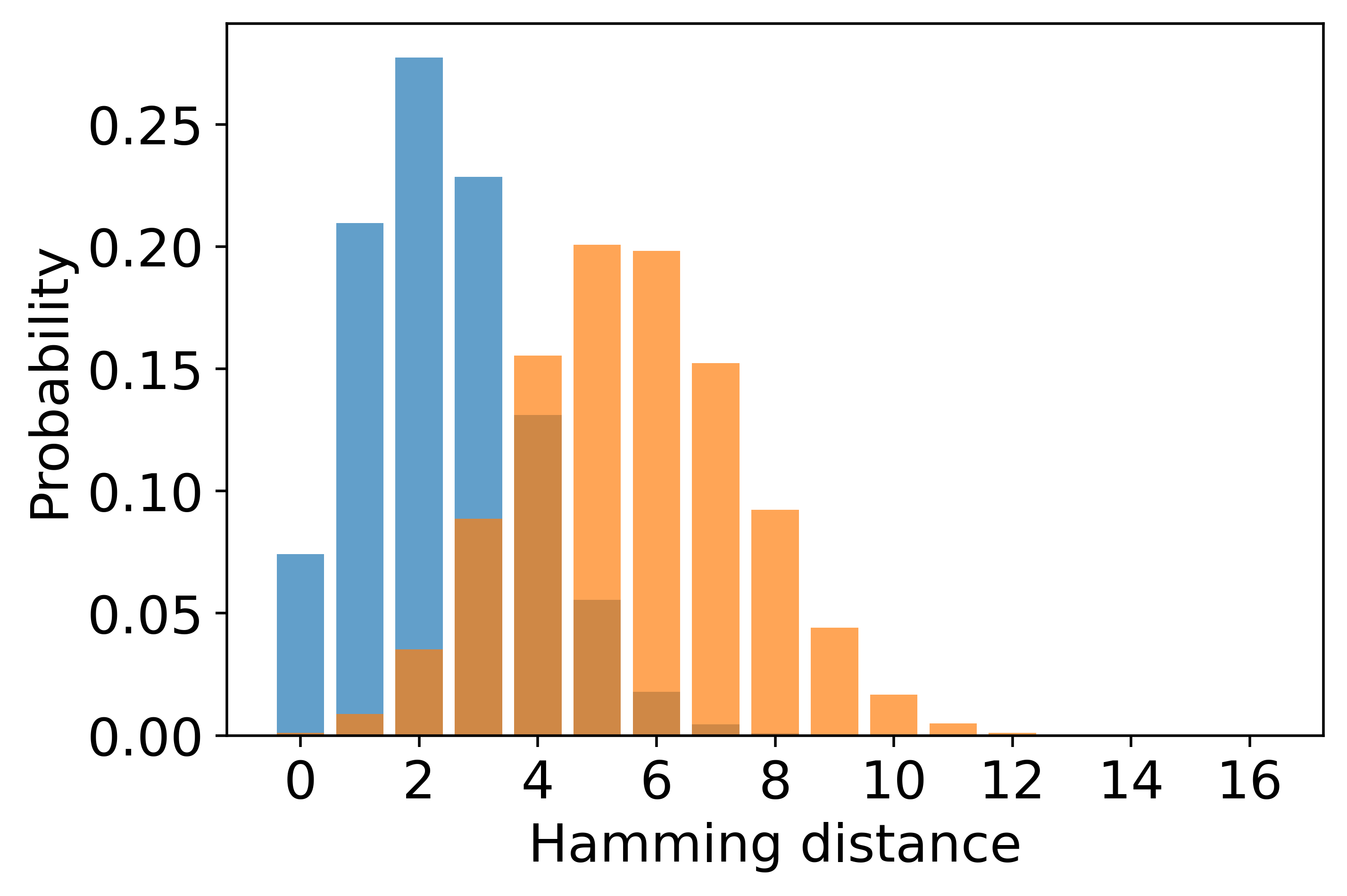}
		\caption{Binary Hash Codes}
		\label{fig:bhc}
	\end{subfigure}
	\begin{subfigure}[b]{0.45\linewidth}
		\centering
		\includegraphics[width=\linewidth]{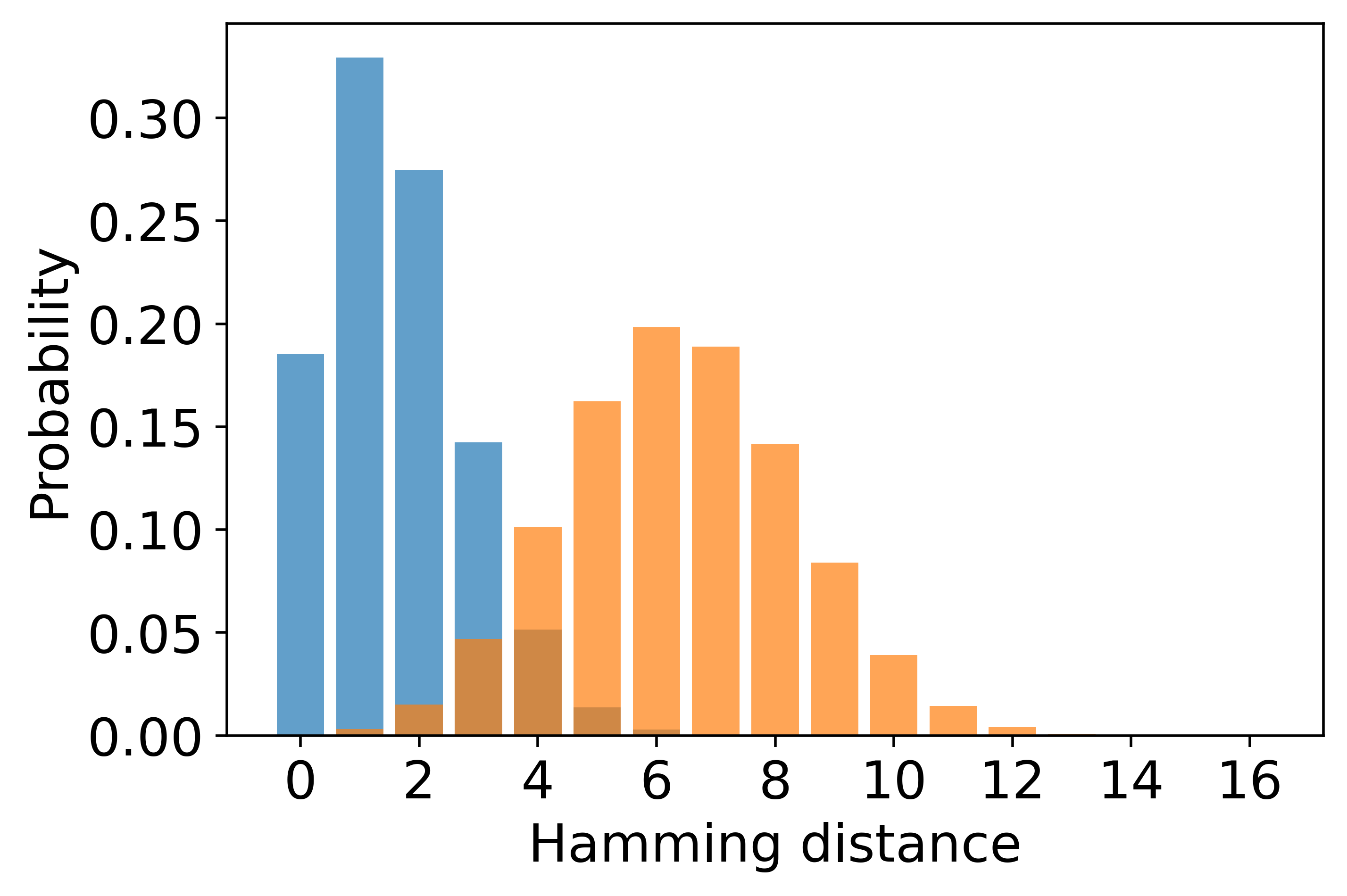}
		\caption{Ternary Hash Codes}
		\label{fig:thc}
	\end{subfigure}
	\caption{A distribution of hamming distance with 16 bits hash codes. Blue and orange colors stand for hamming distance with positive and negative class given a query input.}
	\label{fig:pd}
\end{figure}



From figure \ref{fig:bintwo} to figure \ref{fig:tertwo}, it empirically shown the increased in the expected ternary hamming distance of one bit using the equation from \ref{eq:lukahd} or \ref{eq:kleenhd}. This can then explain the reduction of neighborhood ambiguity in hamming distance between positive and negative classes given a query input. Neighborhood ambiguity is defined as the overlap between the distribution of hamming distance with positive and negative class given a query input \cite{MI_Hash_PAMI19}. If there is no overlap between these two distributions (for one bit, the expected hamming distance = 1), then we can always retrieve the positive images properly. However, the overlapping between the two distributions gave rise in the retrieval error. As shown in the Figure \ref{fig:thc}, ternary hash codes reduced the overlapping (positive classes move towards left and negative classes move towards right), by increasing the expectation of ternary hamming distance after setting some small values within the thresholds as UNKNOWN state (i.e. 0).


{
\bibliographystyle{ieee_fullname}
\bibliography{./bib/Learn2Hash}
}